\documentclass{article}

\usepackage[final]{corl_2020_arxiv} 

\usepackage{grffile}
\usepackage{pratik}
\usepackage{times}
\def \xall {x^{\trm{all}}}
\def \uall {u^{\trm{all}}}

\def \xe {x}
\def \oe {o}
\def \be {\b}
\def \ue {u}
\def \re {r}

\def \qe {q}
\def \pie {{\pi}}

\def \qthe {{q_\th}}
\def \qthie {{q_{\th^i}}}
\def \qthje {{q_{\th^j}}}

\def \qthlage {{q_{\th_{\trm{lag}}}}}
\def \qthilage {{q_{\th^i_{\trm{lag}}}}}

\def \xme {x^{-e}}

\def \bme {\b^{-e}}
\def \pime {{\pi^{-e}}}

\usepackage[linesnumbered,ruled,vlined]{algorithm2e}
\SetAlgorithmName{Algorithm}{Algorithm}{Algorithm}

\SetCommentSty{mycommfont}
\usepackage{multirow}
\usepackage{xspace}
\usepackage{subcaption}
\usepackage{graphicx}
\usepackage{cleveref}
\crefname{algocf}{Algorithm}{Algorithms}

\usepackage{microtype}

\usepackage[small,compact]{titlesec}

\graphicspath{{./pic/}}

\def \alg {MIDAS\xspace}
\def \mytitle {\alg:
Multi-agent Interaction-aware Decision-making with Adaptive Strategies\\
for Urban Autonomous Navigation
}

\title{
\mytitle
}

\author{
  Xiaoyi Chen\\
  Nuro, Inc.\footnotemark\\
  \texttt{xiaoyich@alumni.upenn.edu} \\
   \And
   Pratik Chaudhari \\
   Department of Electrical and Systems Engineering,\\
   University of Pennsylvania.\\
   \texttt{pratikac@seas.upenn.edu} \\
}

\begin{document}
\renewcommand{\thefootnote}{\fnsymbol{footnote}}
\maketitle
\footnotetext{Work done while at the University of Pennsylvania.}
\renewcommand{\thefootnote}{\arabic{footnote}}
\addtocounter{footnote}{-1}

\begin{abstract}
Autonomous navigation in crowded, complex urban environments requires interacting with  other agents on the road. A common solution to this problem is to use a prediction model to guess the likely future actions of other agents. While this is reasonable, it leads to overly conservative plans because it does not explicitly model the mutual influence of the actions of interacting agents. This paper builds a reinforcement learning-based method named MIDAS where an ego-agent learns to affect the control actions of other cars in urban driving scenarios. MIDAS uses an attention-mechanism to handle an arbitrary number of other agents and includes a ``driver-type'' parameter to learn a single policy that works across different planning objectives. We build a simulation environment that enables diverse interaction experiments with a large number of agents and methods for quantitatively studying the safety, efficiency, and interaction among vehicles. MIDAS is validated using extensive experiments and we show that it (i) can work across different road geometries, (ii) results in an adaptive ego policy that can be tuned easily to satisfy performance criteria such as aggressive or cautious driving, (iii) is robust to changes in the driving policies of external agents, and (iv) is more efficient and safer than existing approaches to interaction-aware decision-making.
\end{abstract}



\section{Introduction}

Consider an autonomous vehicle (henceforth called the ``ego'') that is turning left at an intersection:
when
sharing the road with other vehicles, a typical planning
algorithm would predict the forward motion of the other cars and
ego would stop until
it is deemed safe and legal to proceed.
This is a reasonable approach for situations where the right-of-way
is clear. It is however quite
inefficient when the right-of-way is unclear, ego is overly
conservative~\cite{trautman2015robot}, or when human-driven vehicles exploit
a conservative ego policy\footnote{
\href{https://medium.com/backchannel/
the-view-from-the-front-seat-of-the-google-self-driving-car-46fc9f3e6088}{
The View from the Front Seat of the Google Self-Driving Car}}.
We may mitigate this inefficiency by enabling communication between the vehicles.
Perhaps the most natural form of communication comes from mimicking the typical
human behavior of inching forward towards the intersection to make their intent
known to others; this is powerful because it does not require additional
infrastructure and enables other human drivers to reason about
autonomous vehicles the same way they would reason about any other
human-driven vehicle.
This paper devises such driving strategies for autonomous vehicles
that can influence the actions of other drivers,
specifically other human drivers.

The challenges to solving the above problem and our contributions to mitigating them
are as follows.
\begin{enumerate}[noitemsep, topsep=0pt]
    \item \tbf{Driving safely and efficiently is hard because the intent of the other
    drivers is unknown.}
    We model the interaction problem as a partially-observable Markov decision
    process where the ego agent only observes the states of other agents in its vicinity
    and maximizes a reward
    that encourages it to reach a goal region in minimal time while avoiding collisions.
    Non-ego agents drive using an Oracle policy that has full access to trajectories of
    nearby agents.
    A ``driver-type'' variable, which is not observable to others,
    allows the user to tune the policy to be aggressive or cautious.
    \item \tbf{Ability to handle an arbitrary number of
    agents, who may not affect the optimal action.}
    We parametrize the ego policy using an attention-based architecture
    which can handle
    an unordered, arbitrary number of agents in the observation range. Attention
    allows the ego policy to only focus on agents that are relevant to
    decision making.
    In contrast, most current literature
    studies interaction among a fixed~\cite{schmerling2018multimodal} or limited number of agents~\cite{rhinehart2019precog, li2019urban}.
    \item \tbf{Learning interaction policies without knowing the dynamics of
    external agents.}
    We use off-policy reinforcement learning (RL) methods to learn the policy and
    include two variations of typical
    implementations which reduce the variance of the action selection and stabilize the target
    function in temporal-difference learning.
    \item \tbf{A platform for extensive and systematic interaction experiments.}
    We build a simulation environment that enables carefully designed interaction
    experiments with
    a large number of agents in diverse road geometries such as turns, T-intersections, and
    roundabouts. We devise a number of metrics that allow fine-grained
    reporting of safety and efficiency of ego's policy. We can quantitatively
    study how ego influences the actions of external agents in this environment.
    We perform extensive experiments in realistic scenarios to show that MIDAS
    (i) can work across different road geometries,
    (ii) results in an adaptive ego policy that can be tuned easily to satisfy
    performance criteria such as aggressive or cautious driving,
    (iii) is robust to changes in the driving policies of
    external agents, and
    (iv) is more efficient and safe than existing approaches to interaction-aware decision-making.
\end{enumerate}



\section{Problem Formulation}
\label{sec:problem_formulation}

Consider $n$ agents with the state of the $k^{\trm{th}}$ vehicle at time $t \in
\mathbb{Z}_{\geq
0}$ denoted by $x^k_t \in \reals^d$. Denote the combined state of all agents by
$\xall_t := \sbr{x^1_t, \ldots, x^n_t}$.
The control input for the $k^{\trm{th}}$ agent and the control input of the combined system are denoted by $u^k_t$ and $\uall_t = \sbr{\ue_t, \ldots, u^n_t}$ respectively. We will consider a deterministic discrete-time system $\xall_{t+1} = f(\xall_t, \uall_t);\ \xall_0 \sim p_0$
where the initial state $\xall_0$ is drawn from a distribution $p_0$.
The ego agent has index $1$ and to keep the exposition clear, we will simply denote its state, control and driver-type by $\xe_t$, $\ue_t$ and $\be$ respectively. The notation $\xme_t$ and $\bme$ refer to the combined state and driver-type of all agents other than ego.

\heading{Driver-type and observation model} Each agent possesses a
real-valued parameter $\b^k \in [-1,1]$ that models its ``driver-type''.
A large value of $\b^k$ indicates an
aggressive agent and a small value of $\b^k$ indicates that the agent is
inclined to wait for others around it before making progress. The crux of our problem formulation is that an agent cannot observe the driver-type of the other agents.
At each time $t$, each agent $k$ has
access to observations $o^k_t = \cbr{x^i_t:\ d(x^k_t,
x^i_t) \leq D, i=1,\ldots, n}$ that consist of the states of all
agents (including $k$) within some distance $D$.
This model is a multi-agent Partially-Observable Markov
Decision Process (POMDP): agents do not have access to the
entire state of the problem due to a limited observation range, and they
cannot observe the driver-type of others.
Agents-specific goal locations $x_g^k$ are
sampled randomly using a goal distribution $p_g$. The
reward function $r^k(\xall_t; x_g^k, \b^k)$ encourages agents to
reach their goal location in
minimal time while avoiding collisions with other agents.
It encourages different behavior based on $\b^k$.
Ego's goal state is denoted by $x_g$.
Collision
is defined as two agents coming within some
distance $\delta_{\trm{collision}}$ of each other, assuming a
uniform rectangular geometry for all agents.
The deterministic control policy for each agent is $\pi^k(o^k_t;\ \b^k)$.
Given policies of the other agents $\pi^{-e}$, ego maximizes the
objective
\beq{
    \E_{\xall_0 \sim p_0, \xall_g \sim p_g, \bme} \Big[\sum_
    {t=0}^\infty \g^t \re(\xall_t; \xe_g, \b)\ \big|\
    x_{t+1} = f(x_t, u_t), \ue_t = \pie(\oe_t; \be), x_g, x_0, \pi^{-e}\Big].
    \label{eq:objective}
}
to obtain a policy $\pie^*(\be)$ which is a function of its driver-type $\be$.
Ego's policy can thus be
easily evaluated for a different value of $\be$.
The constant $\g < 1$ is a discount factor.

\heading{Simplifications}
We are interested in $n \in [0,25]$ agents and the problem
formulation above is a large decentralized POMDP which is known to be
intractable~\cite{oliehoek2016concise}. We make the following simplifications.
\begin{enumerate}[label=(\roman*), noitemsep, topsep=0pt]
\item We include waypoints as a part of the observation of vector $o^k_t$.
This enables the policy $\pi^k$ to focus the learning only on handling the
interactions instead of spending sample complexity and model capacity on
learning motion-planning behaviors.
A shortest-path algorithm that finds the trajectory from the agent state $x^k_t$ to its goal state $x^k_g$ \emph{without considering other agents on the road as obstacles} is used to compute these waypoints.

\item Control actions of all agents are $u^k_t \in \cbr{0,1}$ which correspond to stop and go respectively.

\item All non-ego agents use the same policy $\pime$. This policy, called the Oracle,
is fixed to a user-designed policy (details in~\cref{ss:evaluation}) and serves as
a benchmark for learned policies.
\end{enumerate}
These simplifications reduce our problem formulation to that of a standard POMDP, albeit with unknown dynamics, where only ego's policy is learned.

\section{The MIDAS Approach}
\label{sec:approach}



\subsection{Off-policy Training}
\label{ss:off_policy}

Define the action-value function
\beq{
    \qe(\oe, \ue; \b) = \E_{\xall_g \sim p_g, \bme} \sbr{\sum_
    {t=0}^\infty \g^t \re(\xall_t;\ \xe_g, \b)\ \big|\
    \xe_0 = x, \ue_0 = \ue, \ue_t = \pie(\oe_t; \b)}
    \label{eq:q}
}
as the expected reward obtained using the policy $\pie$ after starting from
the initial state $x$ and control $u$. Off-policy methods are a popular
technique to estimate this value function and learn the optimal policy by
minimizing the Bellman error, also called the 1-step temporal difference (TD) error,
\beq{
    \trm{TD} :=
    \qe(\oe_t, \ue_t) - \re(\xall_t; \xe_g, \be) - \g\ \qe(\oe_{t+1},
        \pie(\oe_{t+1}; \be))
    \label{eq:td}
}
over a dataset $\DD = \cbr{(\oe_t, \ue_t, \re_t, \oe_{t+1})}_{t=0,\ldots,}$
that
is collected using a (behavior) policy that may
be different from $\pie$. If the value
function $\qthe$
is parameterized
using a
function approximator, say a deep neural network with parameters $\th$, the
TD-error is a function of these parameters $\th$. Further, if the
control-space is discrete (as is the case for us) we can set $\pie(\oe_t; \be) := \argmax_{u'} \qthe(\oe_t, u')$
and learn the optimal value function by performing stochastic gradient descent
to solve
\beq{
    \th^* = \argmin_\th \f{1}{\abr{\DD}} \sum_{(\oe_t,\ue_t,\re_t,\oe_{t+1}) \in \DD} \trm{TD}^2(\th).
    \label{eq:td2}
}
This objective can be seen as the regression error of the first term $\qthe
(\oe_t, \ue_t)$ in~\cref{eq:td} against the sum of the other two terms which are together
called the \emph{target}.
This objective forms the basis for the well-known DQN algorithm~\cite{mnih2015human}.
Off-policy methods are superior to others such as policy-gradient methods
because they reuse the dataset $\DD$ multiple times while learning $\qthe$.
However, implementing them successfully in practice
comes with a few caveats which we describe next.

\heading{Tricks of the trade in off-policy RL}
The $\trm{TD}^2$ objective in~\cref{eq:td2} can be zero even if the value
function is not accurate because the Bellman operator is only a contraction in
the $\ell_\infty$ norm, not the $\ell_2$ norm~\cite{bertsekas2012weighted}.
This leads to instabilities during training which are mitigated by a number of tricks that replace the value function $\qthe(\oe_{t+1}, \argmax_{u'} \qthe(\oe_{t+1}, u'))$ used in~\cref{eq:td}.
A popular scheme is to use time-lagged version of the parameters, 
i.e., use $\max_{u'} \qthlage(\oe_{t+1},  u')$~\cite{mnih2015human}.
Over-estimation bias of the value function estimate~\cite{sutton2018reinforcement}
is countered by using two (or more) copies of the parameters
$\cbr{\qthie, i=1,2}$
along with time-lagged versions for both and using the
minimum of these two
$\min_{i=1,2} \cbr{\qthilage(\oe_{t+1},\argmax_{u'} \qthilage(\oe_{t+1}, u'))}$
to compute the target.
Parameters of both copies are updated using the objective in~\cref{eq:td2}.
A variant of this is Double Q Learning~\cite{hasselt2010double} which uses
$\qthlage(\oe_{t+1}, \argmax_{u'} \qthe(\oe_{t+1}, u'))$.
Notice that the target is computed using the time-lagged parameters but action is computed using the \emph{current, non-lagged} parameters. This ensures a fair evaluation of the
value of the greedy policy given by the \emph{current} parameters.

\heading{Our improvements to off-policy RL}
The above techniques are typically combined together in
state-of-the-art off-policy algorithms. We introduce two variants which further
improve the stability of training. First is to mix the two parameter copies
when using the Double Q Learning trick:
\beq{
    \qthe(\oe_{t+1}, \argmax_{u'} \qthe(\oe_{t+1}, u')) := \min_{i,j \in \cbr{1,2}, i\neq j}
    \cbr{
    \qthilage(\oe_{t+1},
    \argmax_{u'} \qthje(\oe_{t+1}, u'))}.
    \label{eq:twindqn_ours}
}
This forces the first copy, via its time-lagged parameters to be the evaluator for
the second copy and vice-versa; it leads to further
variance reduction of the target in the TD objective.
A second variation we use is to pick the action during \emph{policy evaluation}
using the average of the two copies
\beq{
    \pie(\oe_t; \be) = \argmax_{u'} \f{1}{2} \sum_{i=1}^2 \qthie(\oe_t, u').
    \label{eq:policy_average}
}
Observe that doing so does not invalidate the Bellman equation: if both the
value function estimates are optimal, the action predicted by their
average is also optimal.
The average reduces the variance of picking the best action.
These variations can be implemented with less than 10 lines of
code in state-of-the-art off-policy RL algorithms and only change
the computational cost marginally.

\subsection{Attention-based Policy Architecture}
\label{ss:attention}

Our value function should be \emph{permutation invariant}:
ordering of the states in the
observation vector $\oe_t$ should not affect ego's action.
It should also be able to \emph{handle an arbitrary number of other agents}.
The optimal action only depends on the \emph{actionable information}
~\cite{soatto2013actionable}, namely agents
whose states cause a change in ego's action.
We next discuss an architecture that bakes in these properties.

\heading{Permutation-invariant input representation}
\cite{zaheer2017deep} shows that a function
$\phi(A)$ operating on
a set $A$ is invariant to permutation of the elements in $A$ iff it can be
decomposed as $\phi(A) \equiv \rho(\sum_{a \in A} \varphi(a))$ for some
functions $\varphi$ and $\rho$. This is a remarkable result because
both $\varphi, \rho$ can now be learnt to build invariance.
The summation (average pooling) enables $\phi(A)$ to handle
sets with varying number of elements. Observe
however that the
summation assigns the same weight to all elements in the input.
As we see in our experiments, a value
function using this ``DeepSet'' architecture
is likely to be distracted by agents that do not inform the optimal action.

\heading{Attention}
An attention mechanism~\cite{Vaswani2017AttentionIA} takes in elements of the set $A$ and
outputs a linear combination
$z(a) = \sum_{a' \in A} \a_{a,a'}\ \varphi_v(a')$
where
$\varphi_v(a)$ is the ``value'' embedding of $a$ and weights $\a_{a,a'} =
\inner{\varphi(a)}{\varphi(a')}$ compute the similarity of different elements.
are similar.
A generalization of this is the attention module of~\cite{Vaswani2017AttentionIA} which uses $\a_{a,a'} =
\inner{\varphi_k(a)}{\varphi_q(a')}$. Similarity is thus
computed across a set of ``keys'' $\varphi_k$ and a set of ``queries'' $\varphi_q$, all of which are learned.
An attention module is an elegant way for
the value function to learn key, query, value embeddings that pay more attention
to parts of the input that are more relevant to the output. E.g,~\cite{leurent2019social}
sets the
key to be an encoding of ego's state, which is compared to query and value embeddings
generated from the observation vector.

\heading{Set-based attention~\cite{lee2018set} in MIDAS}
The summation in attention
throws away higher-order correlations since
$\varphi_v$ is computed independently for each $a$.
A better representation for set-valued inputs can be created using:
(i) a \emph{set-attention block} (SAB) (we use its
efficient version \emph{induced} SAB) which builds
$\varphi_v(a)$ itself using self-attention;
(ii) another self-attention mechanism
to aggregate features instead of simple summation;
this is called \emph{pooling by multi-head
attention} (PMA), (iii) regularization techniques such as
layer-normalization~\cite{ba2016layer} and
residual connections~\cite{he2016identity}, and (iv) \emph{multi-head
attention}~\cite{Vaswani2017AttentionIA}
that computes inner products independently across
subspaces of the embedding.

\heading{Encoding ego's driver-type}
We want ego's driver-type information to only affect the encoding of its own
state, not the state of the other agents, in the observation vector; this is akin
to masking NLP~\cite{devlin2018bert}.
We use a two-layer perceptron with ReLU nonlinearities to embed the scalar
variable $\be$ and add the output to the encoding of ego's state.




\subsection{Designing the reward function}
\label{ss:reward_design}

The reward function for ego in our problem is designed to encourage
it to make progress towards its $\xe_g$ while minimizing the time taken and the
collisions with the other agents on the road.
It consists of
a time-penalty for every timestep, reward for non-zero speed, timeout penalty that
discourages ego from stopping the traffic flow
(this includes a stalement penalty where all nearby agents including ego are
standstill waiting for one of them to take initiative and break the tie), a
collision penalty and a penalty for following too close to the agent in front.
All these sub-rewards except
the last one depend linearly on $\be$; they have the form $w \be + b$ where
the weight $w$ and the bias $b$ are chosen (see~\cref{app:reward_coefficients})
to achieve good performance on downstream metrics
such as time-to-finish, collision, timeout and success rates
\emph{using a generic RL agent}.
We emphasize that designing ego's reward this way
is reasonable and indeed what an algorithm designer
will do in practice~\cite{hu2019interaction}.

\section{Experiments}
\label{sec:experiments}

\subsection{Evaluation Methodology}
\label{ss:evaluation}

The outcome of a multi-agent planning problem under partial observations is
inherently noisy. It is therefore important to build an evaluation
suite which provides fine-grained understanding but also allows
studying complex scenarios.

\begin{wrapfigure}{r}{0.2\textwidth}
\centering
\includegraphics[width=0.2\textwidth]{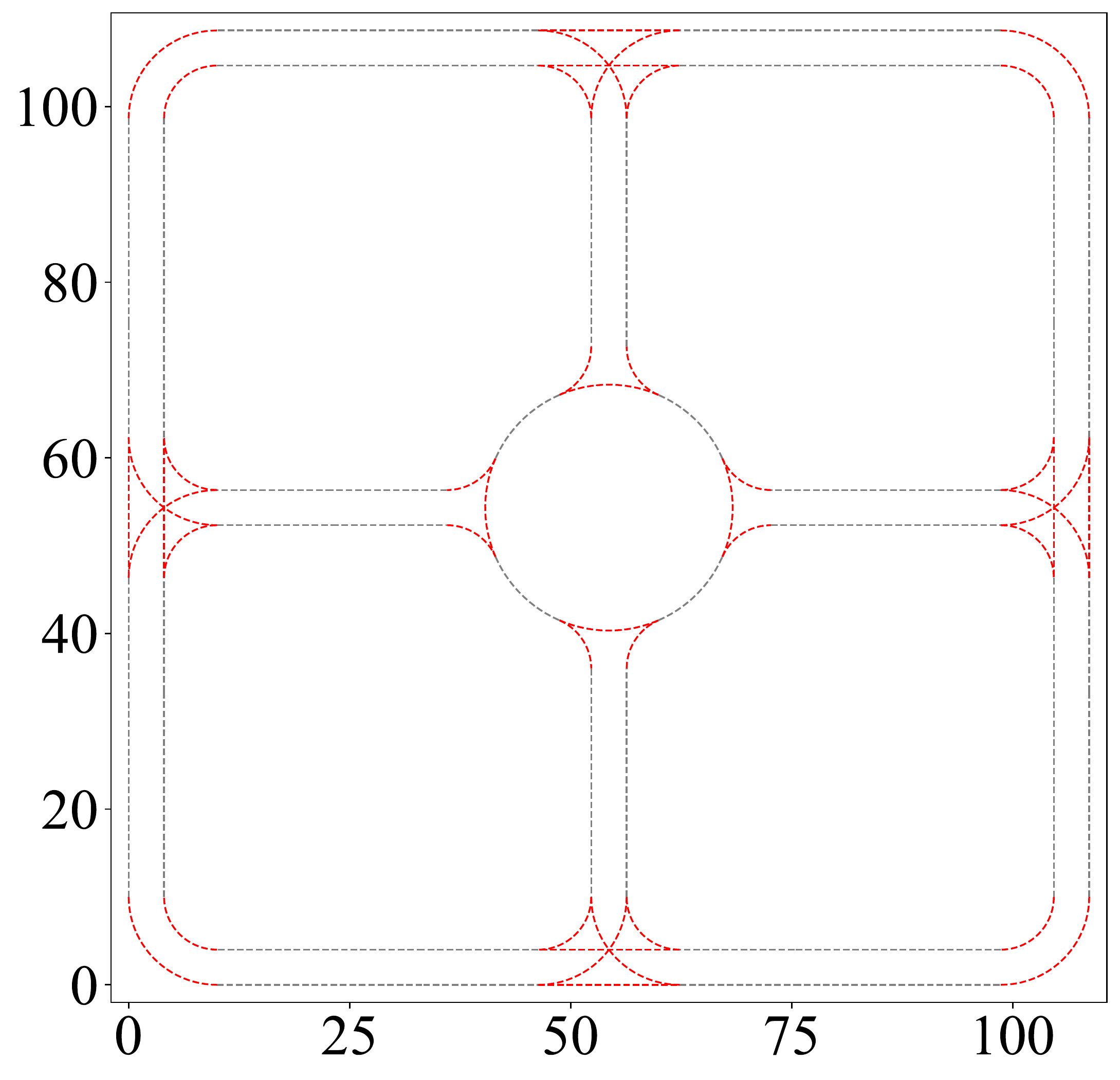}
\caption{\tbf{Map of the environment}}
\label{fig:3-3_neighborhood_map}
\end{wrapfigure}

\heading{Simulation environment}
\cref{fig:3-3_neighborhood_map} shows our simulation environment which consists
of 4 T-intersections, 4 corners and a roundabout.
Lanes with direction information are encoded into the map.
An episode begins with each
agent initialized at a random location and with a randomly chosen
goal location.
Each agent is randomly assigned
a driver type $\beta \in [-1,1]$: the driver-type determines the agent's
velocity as $v = 2.7 \beta
+ 8.3$; these constants are based on
the recommended turning velocity of the turning radii in the
environment~\cite{hancock2013policy}. A timestep in the environment is 0.1s.
For Oracle agents (see below), the observation vector contains the states of all
agents within a travel-distance of 9.2m along their paths.
Non-Oracle agents do not have global information
about other agents' path information. Their
observation vector contains the locations of all agents
within a \emph{Euclidean} distance of 10m and is created in an ego-centric
coordinate frame; this is a simple technique to make the policy generalize trivially to new maps without
further training.


\heading{Oracle planner} Non-ego agents run
an Oracle planner which is executed as follows. For any two agents $i, j, i\neq j$ let
$\tau^{ij}_{0,1}$ be the time-to-collision (TTC) if agent $i$ takes action zero (stops) for
all time-steps henceforth and agent $j$ moves forward along its waypoints for all time-steps henceforth.
The TTC can be infinite. At every timestep, the Oracle planner of $i$ compares
$\tau^{ij}_{0,1}$ with $\tau^{ij}_{1,0}$ for all agents $j$ within its observation range
and stops if the former is larger. A tie between the two values is broken by giving priority to $\min(i,j)$, but any other mechanism can be used here.

\heading{Systematically creating an evaluation dataset}
%
We curate (i) \emph{generic episodes} that
consist of a random number of agents with uniformly random
initial and goal locations, (ii) \emph{collision episodes} which
ensure that
the ego will collide with at least one other agent in the future if it does not
stop at an appropriate timestep, and (iii) \emph{interaction episodes} which
ensure that 2--3 agents (including ego) arrive at a location
simultaneously (or within 4 timesteps of each other so as to trigger our
collision threshold). Interaction episodes are illustrated in~\cref{app:interaction_settings}. Ego cannot do well in interaction episodes
unless it negotiates with other agents.
We randomize over the number of agents,
driver-types, agent IDs, road geometries, and add small
perturbations to their arrival time to construct
1917 interaction episodes.
We use a mix of the three kinds of episodes during training and
use a mix of generic and interaction episodes, reflective of general driving settings with an emphasis on interactions, for validation and testing.
We separately report the test performance of interactive episodes. 
See~\cref{app:train_validation_test_configs} for more details.
Curating the dataset in this fashion aids the reproducibility of results compared
to using random seeds to initialize the environment, as is commonly done.

\heading{Evaluation Metrics}
We evaluate performance based on (i) the \emph{time-to-finish} which is the
average episode length, and (ii) \emph{collision, timeout and success rate}
which refer to the percentage of episodes that end with the
corresponding status (the three add up to 1). To qualitatively
compare performance, we prioritize collision rate (an indicator for safety)
over the timeout rate and time-to-finish (which indicate efficiency).
Performance of the Oracle planner is reported over 4 trials;
performance of the trained policy is reported across 4 random seeds.

\heading{Baselines}
We perform comprehensive benchmarking and compare MIDAS against
the Oracle planner which is the gold-standard and acts as an upper bound on performance,
a simple \emph{Car Follower} that keeps a fixed minimum distance from the agent in front
which acts as a lower bound on performance, and learning-based models such as
a multi-layer perceptron (MLP), Deep Set~\cite{zaheer2017deep} which is a popular
architecture for prediction models~\cite{gupta2018social},
and Social Attention~\cite{leurent2019social} to fit the value function.
\cref{app:model_implementation_details} provides more implementation details.

%
%

\subsection{Results}
\label{ss:results}

\begin{figure*}[!htpb]
\begin{minipage}[t]{0.58\textwidth}
    \begin{figure}[H]
    \centering
    \captionsetup[subfigure]{justification=centering}
    \begin{subfigure}[t]{.49\textwidth}
        \includegraphics[width=\linewidth]{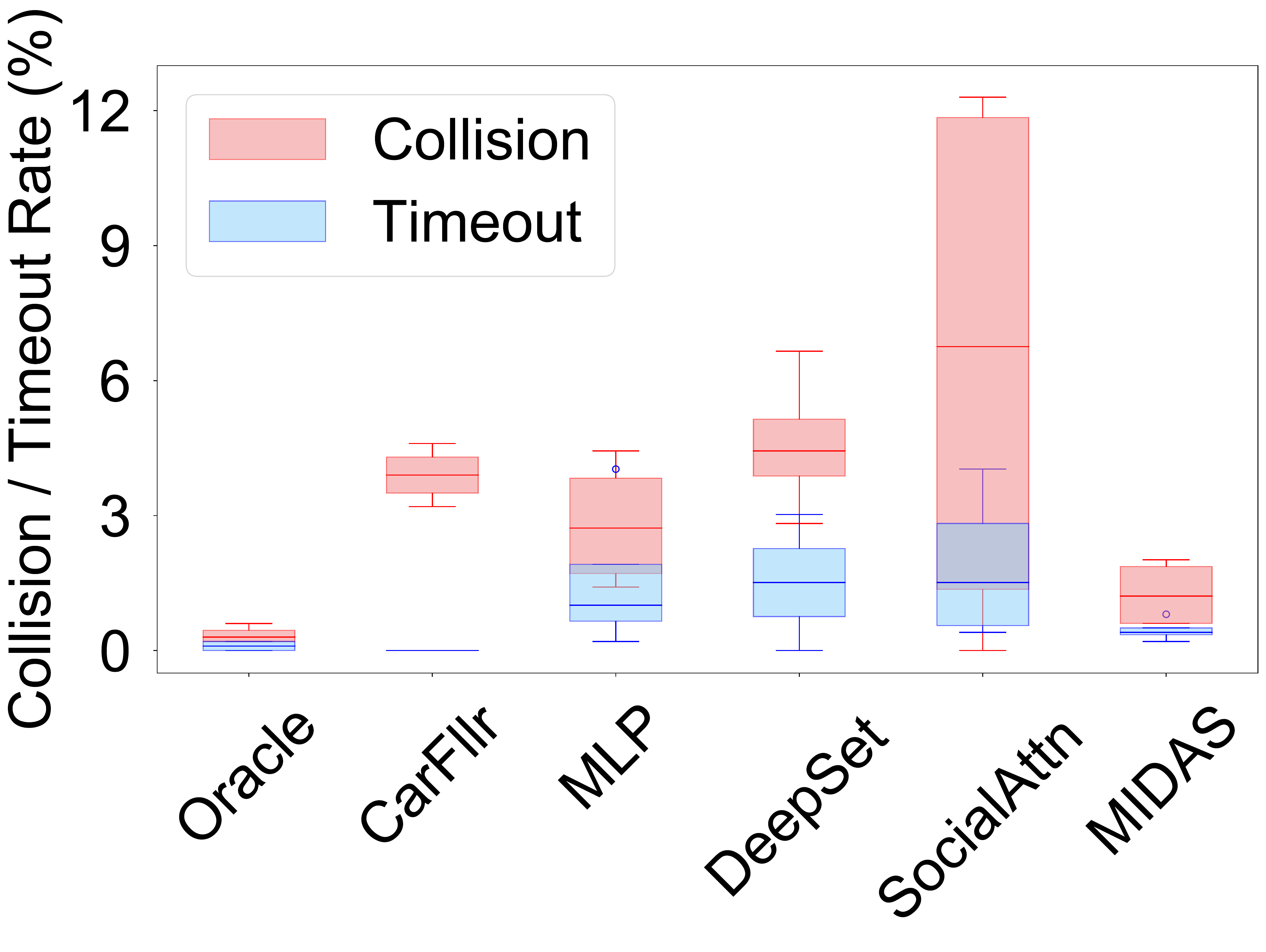}
    \caption{Test set collision and timeout rates}
    \label{fig:box_plot_short_traj_eval_set_cls_to}
    \end{subfigure}
    \begin{subfigure}[t]{.49\textwidth}
        \includegraphics[width=\linewidth]{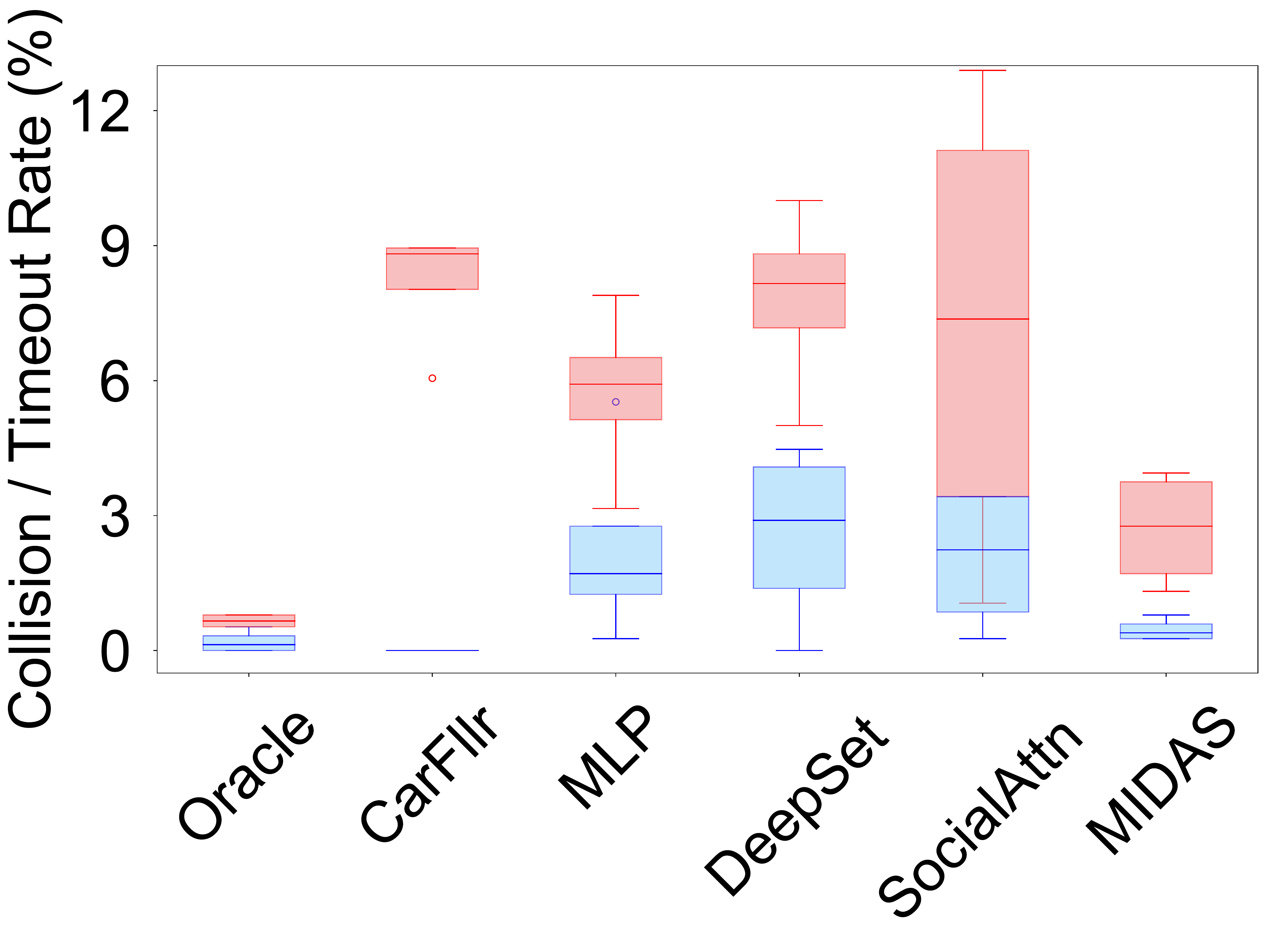}
    \caption{Test interaction set collision and timeout rates}
    \label{fig:box_plot_testing_interaction_set_cls_to}
    \end{subfigure}

    \begin{subfigure}[t]{.49\textwidth}
        \includegraphics[width=\linewidth]{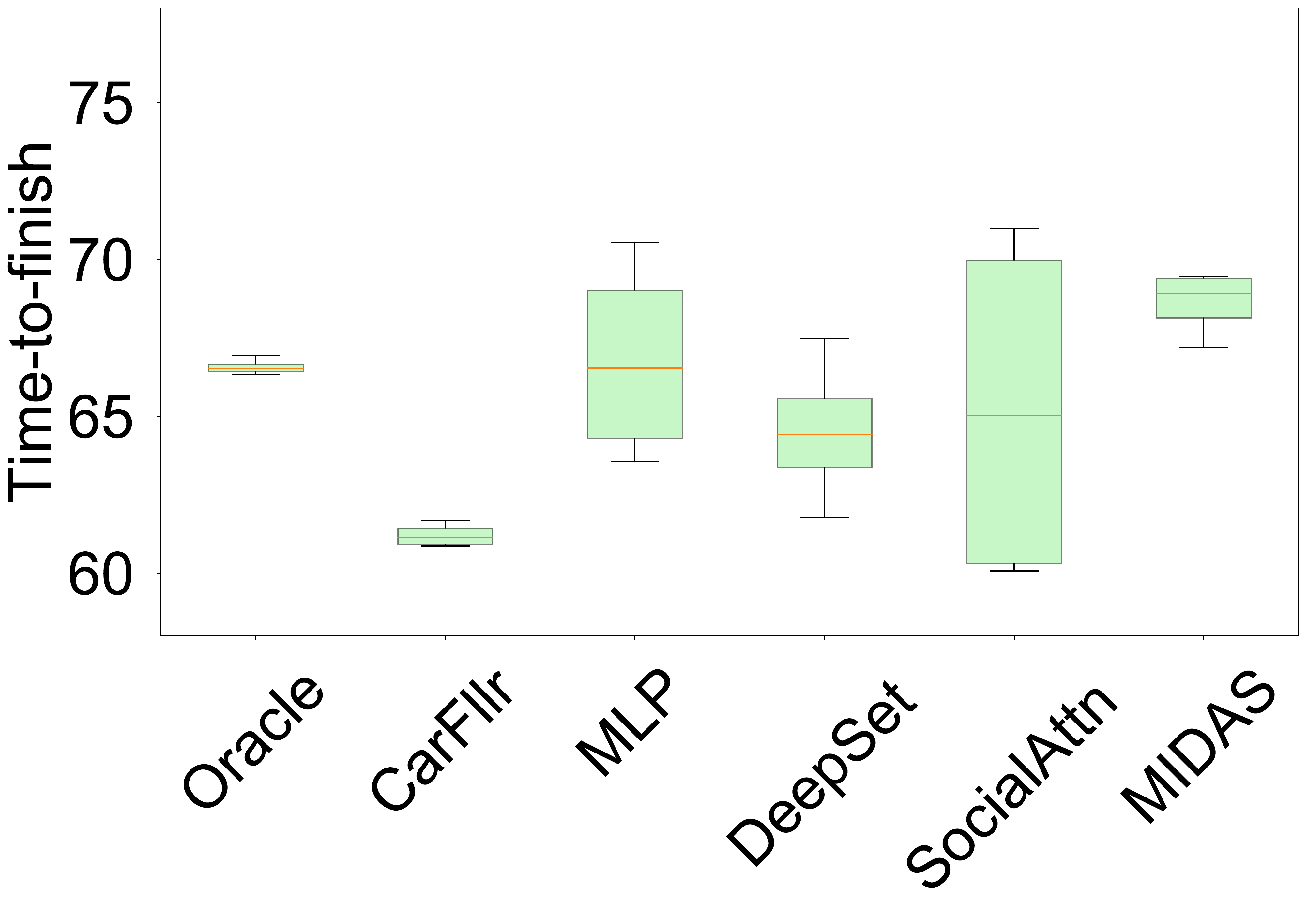}
    \caption{Test set time-to-finish}
    \label{fig:box_plot_short_traj_eval_set_ttf}
    \end{subfigure}
    \begin{subfigure}[t]{.49\textwidth}
        \includegraphics[width=\linewidth]{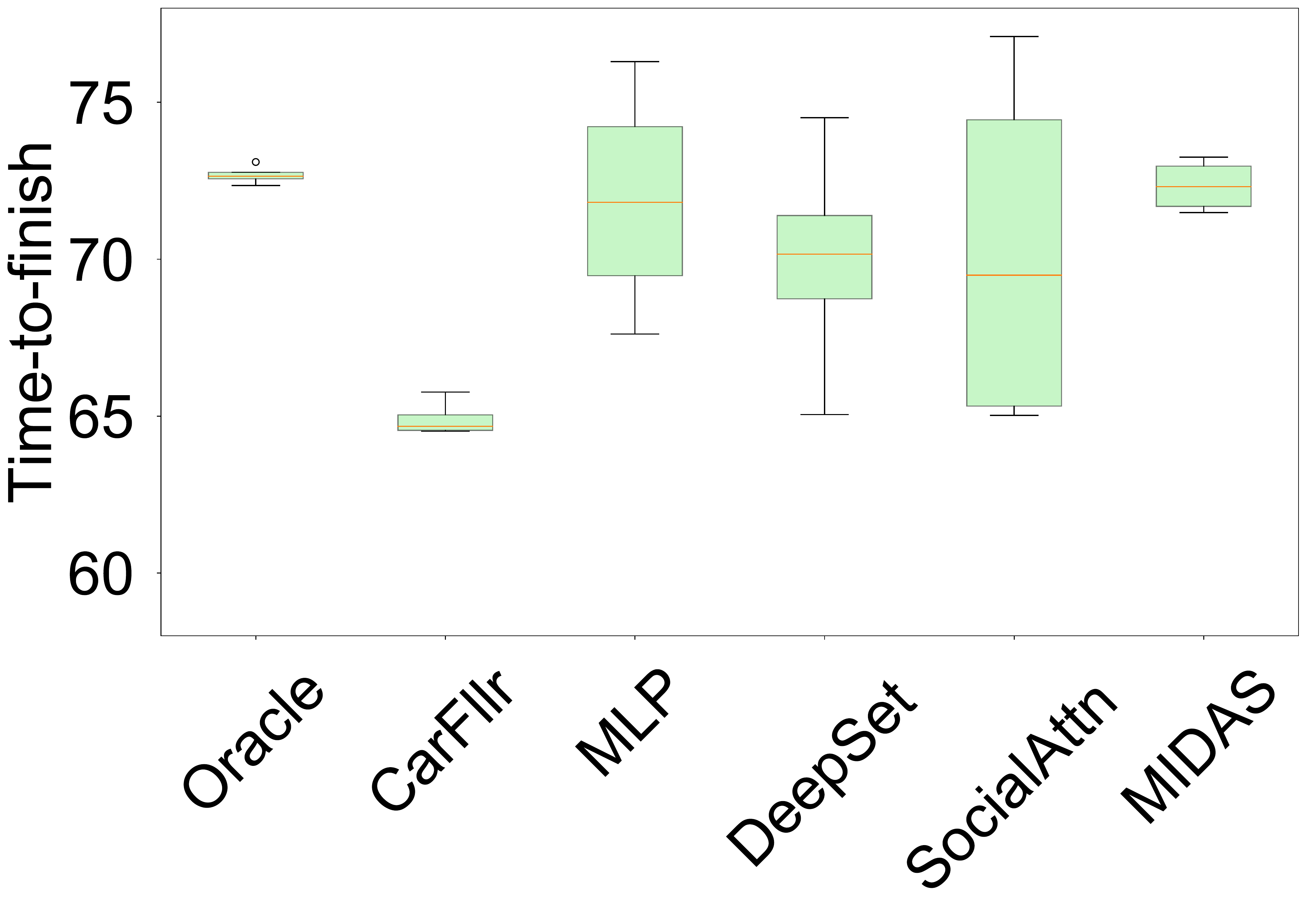}
    \caption{Test interaction set time-to-finish}
    \label{fig:box_plot_testing_interaction_set_ttf}
    \end{subfigure}
    \caption{\tbf{MIDAS is safer than Car-Follower and other learned planners and similar to Oracle in efficiency.}
    Time-to-finish is average episode length within the set. Collision, timeout rates refer to the percentage of episodes that end with the respective status. CarFllr, SocialAttn refer to Car-Follower and Social Attention, respectively.}%
    \label{fig:6-1_pfmc_boxplots}%
    \end{figure}
\end{minipage}
\hspace{1em}
\begin{minipage}[t]{0.41\textwidth}
    \begin{figure}[H]
    \centering
    \captionsetup[subfigure]{justification=centering}
    \begin{subfigure}[t]{.49\textwidth}
        \includegraphics[width=\linewidth]{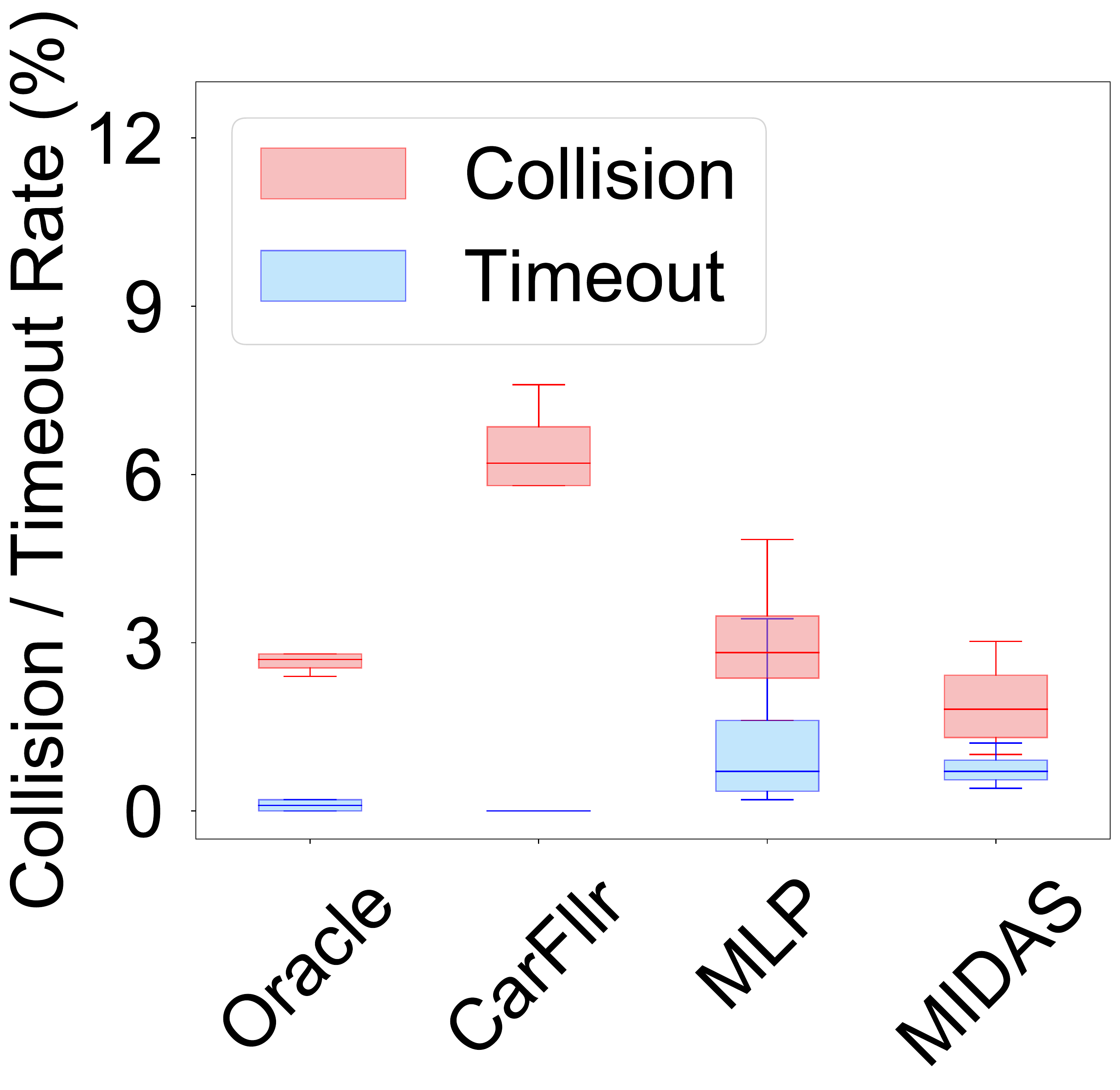}
    \caption{Test set collision and timeout rates}
    \label{}
    \end{subfigure}
    \begin{subfigure}[t]{.49\textwidth}
        \includegraphics[width=\linewidth]{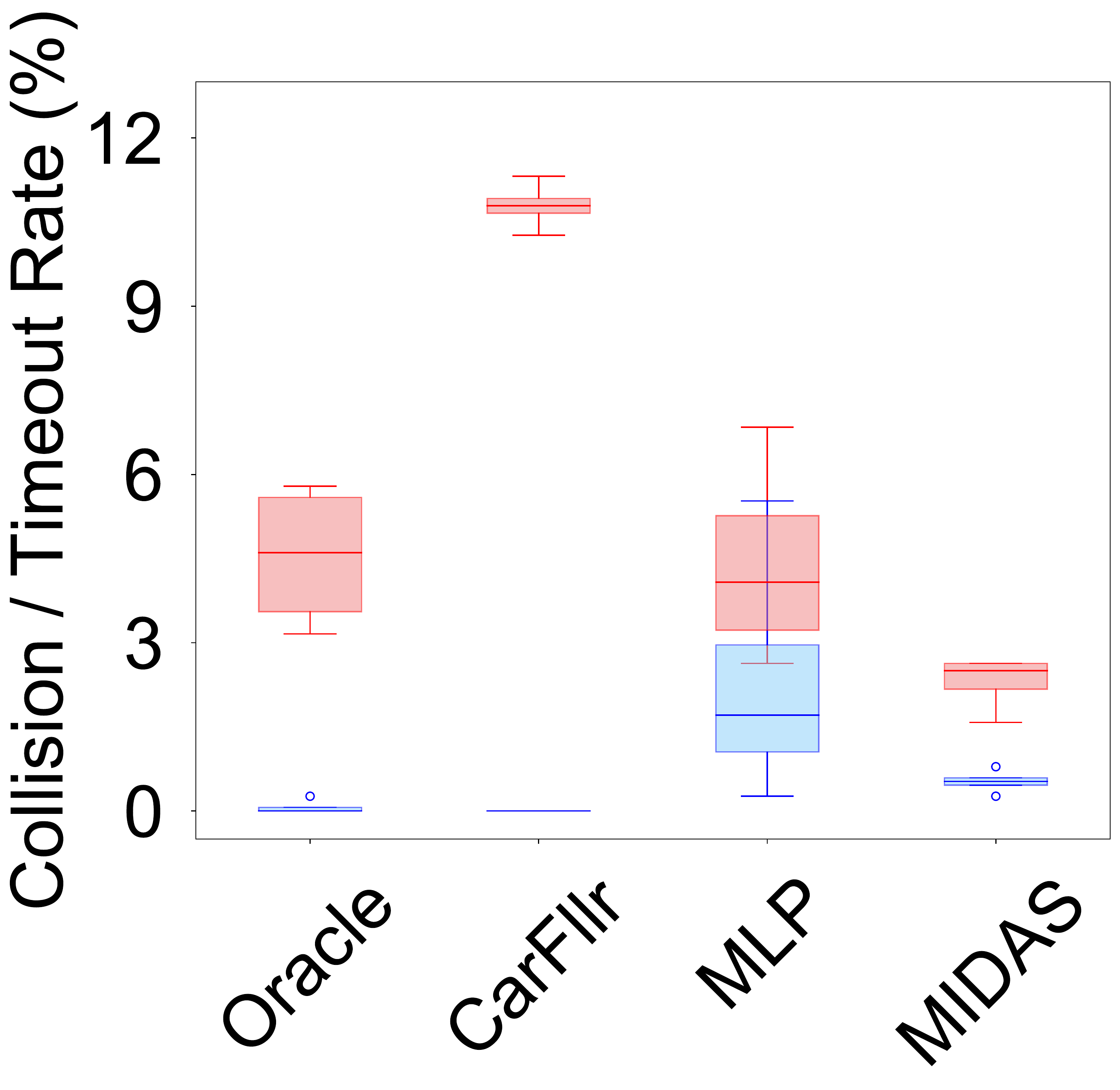}
    \caption{Test interaction set collision and timeout rates}
    \label{}
    \end{subfigure}

    \begin{subfigure}[t]{.49\textwidth}
        \includegraphics[width=\linewidth]{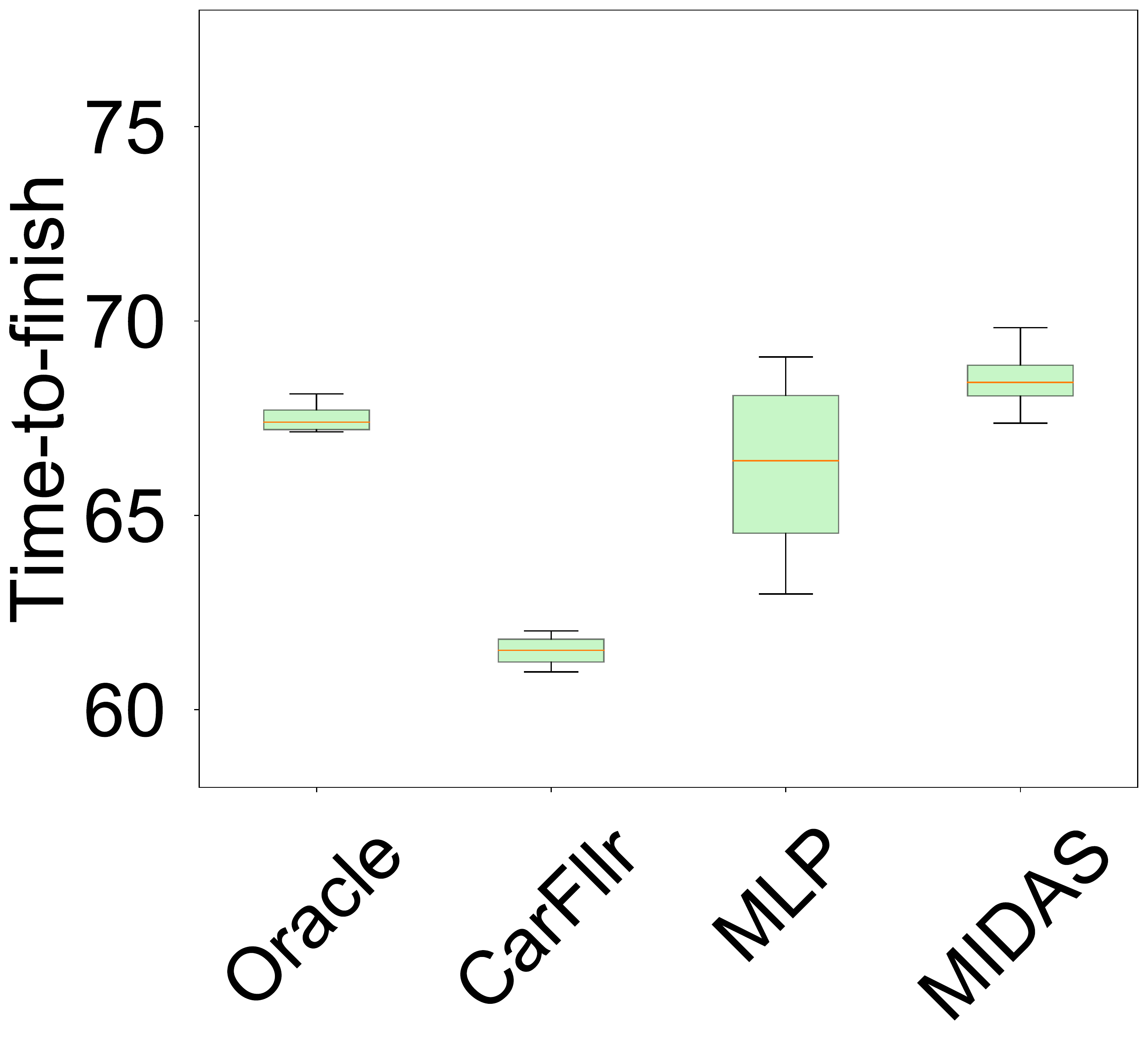}
    \caption{Test set time-to-finish}
    \label{}
    \end{subfigure}
    \begin{subfigure}[t]{.49\textwidth}
        \includegraphics[width=\linewidth]{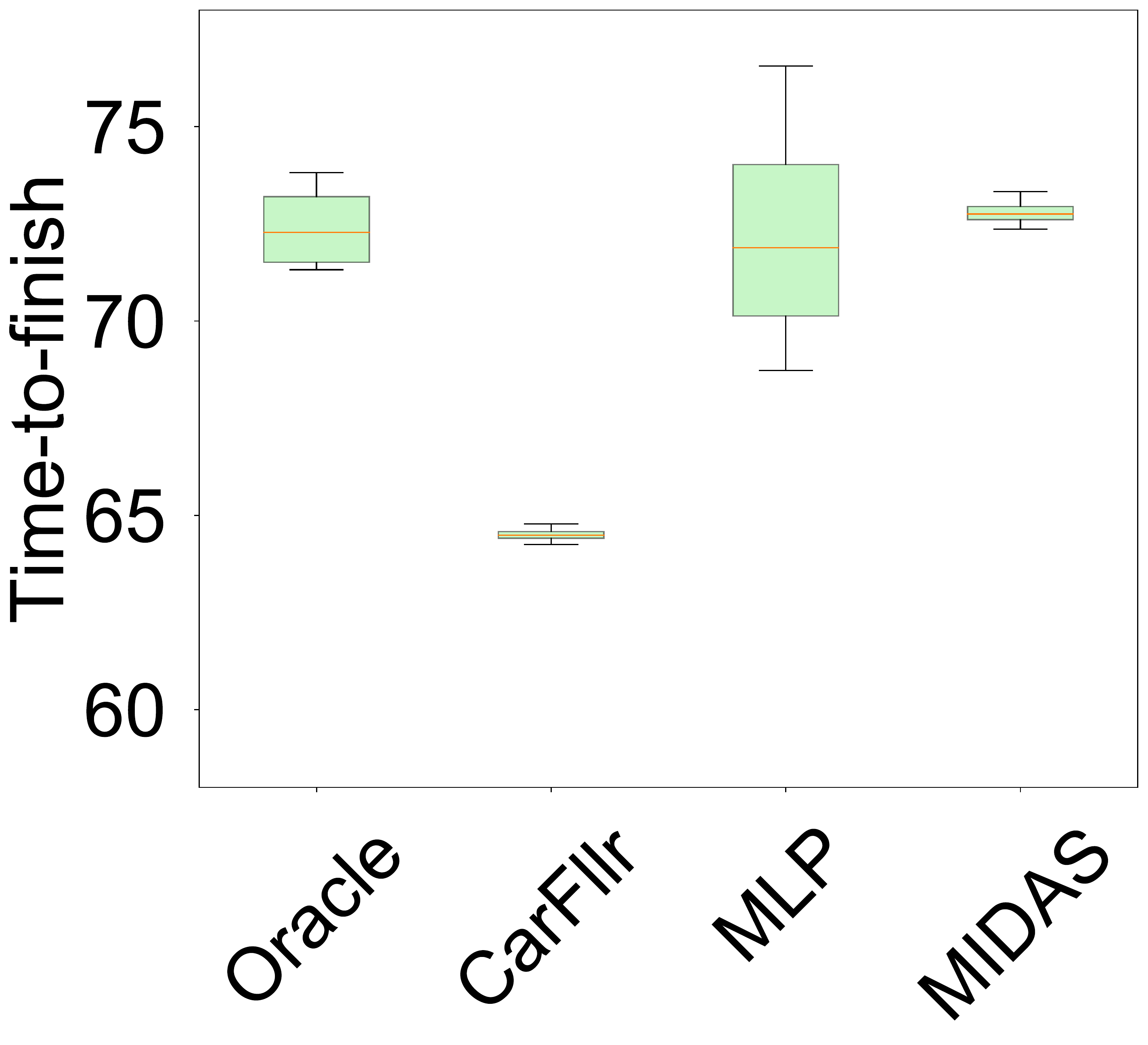}
    \caption{Test interaction set time-to-finish}
    \label{fig:box_plot_testing_interaction_set_ttf_aan=0.1}
    \end{subfigure}
    \caption{\tbf{MIDAS is more generalizable across different driving policies for the other agents.} Test performance with action noise for external agents. Oracle and Car-Follower collision rates greatly increase from those in~\cref{fig:6-1_pfmc_boxplots} while the performance of MLP and MIDAS is unchanged.
    }%
    \label{fig:6-1_pfmc_boxplots_aan}%
    \end{figure}
\end{minipage}
\end{figure*}

\cref{tab:6-1_experiment_results} reports test
performance of all algorithms across three different metrics on the test set
and the test interaction set. Box plots in~\cref{fig:6-1_pfmc_boxplots} show the same information; training curves are given in~\cref{app:training_curves}.
We next discuss these results in the context of specific questions.

\begin{table*}[!t]
\renewcommand{\arraystretch}{1.4}
\caption{\tbf{Summary of empirical results}.
Time-to-finish is
average episode length within the set. Collision, timeout and success rates refer to the percentage of episodes within the set that end with the respective status. Lower collision rate indicates higher safety. Lower timeout rate and time-to-finish indicate higher efficiency. Higher success rate is better.
We verified that most timeout cases for MIDAS are caused by ego
stepping into the TTC thresholds of other agents and then stopping
for other agents, while other Oracle-driven agents also choose to wait for ego,
leading to a stalemate. In reality, stalemates like this are unlikely to happen because
once ego stops, the other drivers would likely move forward and break the tie.
}
\label{tab:6-1_experiment_results}
\begin{center}
\resizebox{\textwidth}{!}{
        \begin{tabular}{|p{2.75cm}|c|c|c|c|c|c|c|c|}
             \toprule
             \multirow{2}{*}{\tbf{Planner}} & \multicolumn{4}{c|}{\tbf{Test Set}} & \multicolumn{4}{c|}{\tbf{Test Interaction Set}}\\
             \cline{2-9}
              & \tbf{Time-to-finish} & \tbf{Collision (\%)} & \tbf{Timeout (\%)} &
            \tbf{Success (\%)} & \tbf{Time-to-finish} & \tbf{Collision (\%)} &
            \tbf{Timeout (\%)} & \tbf{Success (\%)} \\
             \midrule
             Oracle     & 66.57 $\pm$ 0.23 & 0.35 $\pm$ 0.17 & 0.10 $\pm$ 0.10 & 99.55 $\pm$ 0.17
                        & 72.68 $\pm$ 0.27 & 0.66 $\pm$ 0.13 & 0.20 $\pm$ 0.22 & 99.14 $\pm$ 0.11\\
             Car Follower & 61.20 $\pm$ 0.32 & 3.90 $\pm$ 0.54 & 0.00 $\pm$ 0.00 & 96.10 $\pm$ 0.54
                        & 64.91 $\pm$ 0.51 & 8.16 $\pm$ 1.22 & 0.00 $\pm$ 0.00 &
                        91.84 $\pm$ 1.22\\
             \midrule
             MLP        & 66.78 $\pm$ 2.85 & 2.82 $\pm$ 1.25 & 1.56 $\pm$ 1.47 & 95.61 $\pm$ 2.52
                        & 71.88 $\pm$ 3.30 & 5.72 $\pm$ 1.69 & 2.30 $\pm$ 1.95 & 91.97 $\pm$ 2.62\\
             DeepSet~\cite{zaheer2017deep}
                        & \textbf{64.52 $\pm$ 2.04} & 4.59 $\pm$ 1.37 &
             1.51 $\pm$ 1.13 & 93.90 $\pm$ 1.37
                        & \textbf{69.97 $\pm$ 3.35} & 7.83 $\pm$ 1.81 & 2.57 $\pm$ 1.78 & 89.61 $\pm$ 1.15\\
             SocialAttention~\cite{leurent2019social}
             & 65.27 $\pm$ 5.06 & 6.45 $\pm$ 5.59 &
             1.86 $\pm$ 1.48 & 91.68 $\pm$ 4.36
                        & 70.27 $\pm$ 5.21 & 7.17 $\pm$ 4.75 & 2.04 $\pm$ 1.41 & 90.79 $\pm$ 3.53\\
             MIDAS (Ours) & 68.61 $\pm$ 0.92 & \textbf{1.26 $\pm$ 0.66} & \textbf{0.45 $\pm$ 0.22} & \textbf{98.29 $\pm$ 0.54}
                        & 72.34 $\pm$ 0.74 & \textbf{2.70 $\pm$ 1.14} & \textbf{0.46 $\pm$ 0.22} & \textbf{96.84 $\pm$ 1.05}\\
             \bottomrule
        \end{tabular}
}
\end{center}
\vspace*{-2em}
\end{table*}

\tbf{1. Is the learned model safe and efficient?}
%
First, our experiments show that learning interactive driving policies is indeed beneficial: MLP and MIDAS both achieve lower collision rates than the rule-based Car-Follower in both general driving (\cref{fig:box_plot_short_traj_eval_set_cls_to}) and interactive driving (\cref{fig:box_plot_testing_interaction_set_cls_to}) settings.
MIDAS is safer than the naive
rule-based Car-Follower and other attention-based models
while being as efficient as the Oracle.
MIDAS has lower collision rate than all other learned models. We also achieve the lowest timeout rates among all learned models. In terms of time-to-finish (\cref{fig:box_plot_short_traj_eval_set_ttf}, \cref{fig:box_plot_testing_interaction_set_ttf}), it is remarkable that
MIDAS is similar to Oracle even though the former does not have all the
information (long-time horizon, tie-breaking priority) that the Oracle has access to.


\tbf{2. Can the model generalize across different driving policies for the
other agents?}
At test time, we add Bernoulli noise of probability 0.1 to the actions of other agents to model the fact that driving policies of other agents may be different from each other, and different from our gold-standard Oracle.
As shown in~\cref{fig:6-1_pfmc_boxplots_aan}, the Oracle's collision rate is
now larger, which can be directly attributed to conflicting actions taken by
the agents during interaction.
The rule-based Car-Follower \emph{becomes} more aggressive in this case and although its time-to-finish is small, the collision rates are quite high. MLP performs about as well as the Oracle.
In contrast to these, MIDAS has similar collision/timeout rates and time-to-finish from~\cref{fig:6-1_pfmc_boxplots}
and outperforms Oracle in both driving settings (\cref{fig:6-1_pfmc_boxplots_aan}). This shows that
MIDAS can generalize better to different driving policies of other agents
on the road.

\begin{wrapfigure}{r}{0.35 \textwidth}
\centering
\includegraphics[width=\linewidth]{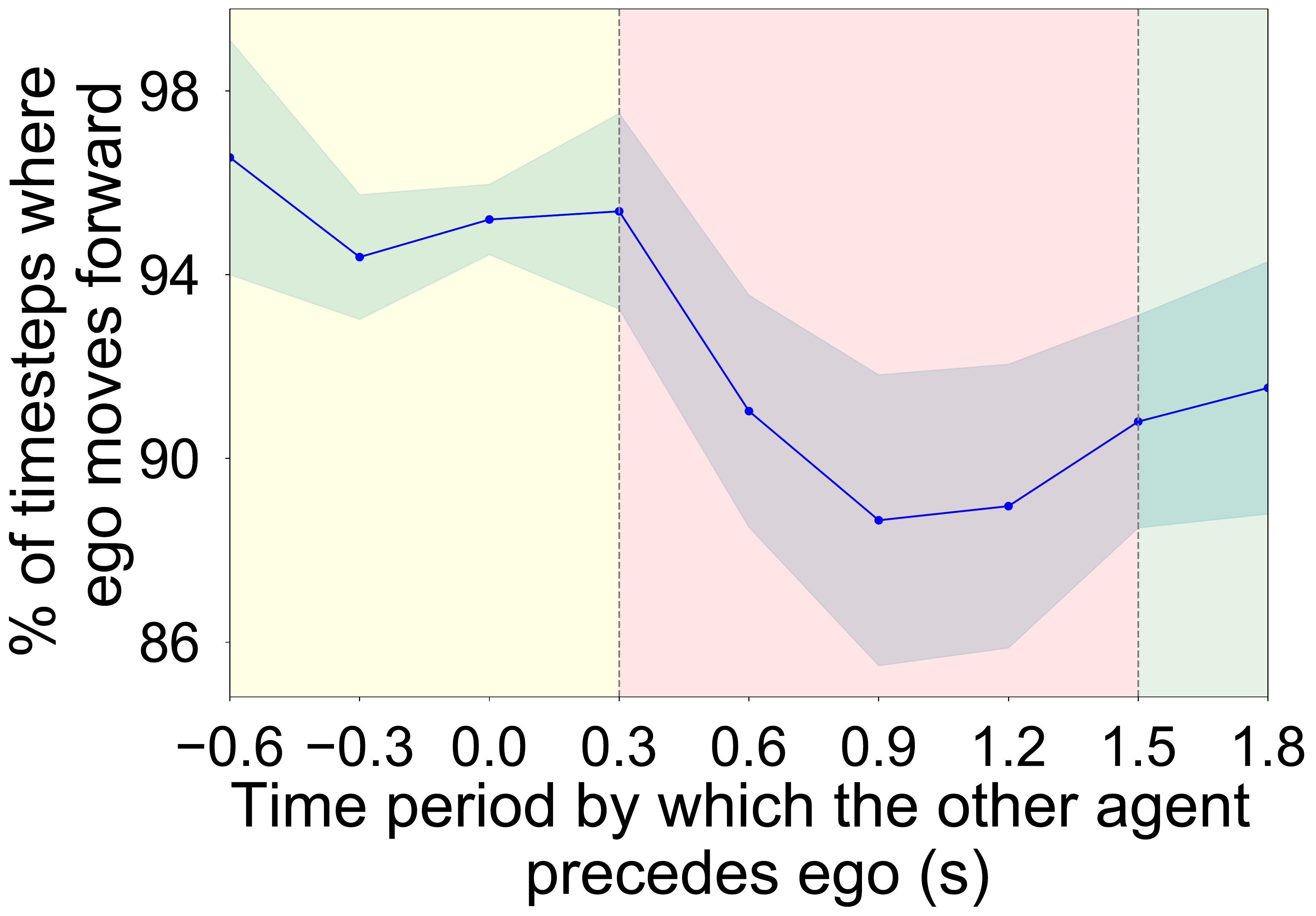}
\caption{In ambiguous situations without clear right-of-way (yellow) or if the
intersection clears before ego arrives (green), MIDAS drives forward. But it stops
more often if a collision is more likely (red).}
\label{fig:midas_bhvr_change_perturbation}
\includegraphics[width=\linewidth]{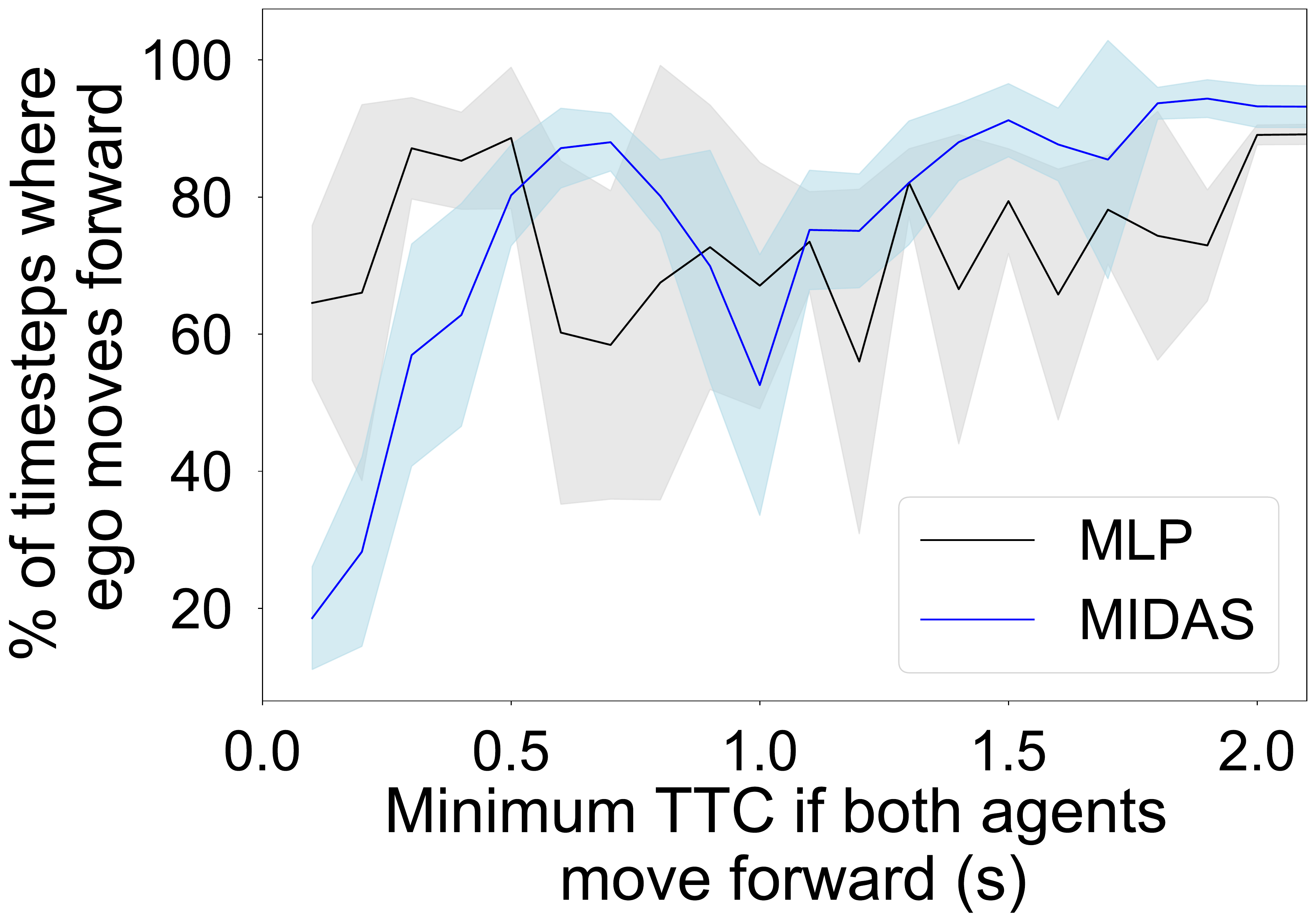}
\caption{For small TTC when collision is immediate, MIDAS is less likely to proceed
than MLP, indicating that
MIDAS does not blindly drive forward in safety critical situations.}
\label{fig:midas_vanilla_min_ttc3_ego_go_pcts}
\vspace*{-1.5em}
\end{wrapfigure}

\tbf{3. Does MIDAS influence the actions of other agents?}
We curate test episodes where ego and one other agent arrive
simultaneously at a given location if they take their nominal actions. The arrival time of the other agent is then perturbed to study how MIDAS changes its actions.
The X-axis in~\cref{fig:midas_bhvr_change_perturbation} is the time period (in seconds)
by which the other agent \emph{precedes} ego. The Y-axis shows the percentage of
timesteps where $\ue_t = 1$, i.e., ego drives forward.
The size of the agents is such that if the arrival times are within 0.3s of
each other, either ego or the other agent could go first. We see that ego
takes the initiative in more than 94\% of the cases if it arrives earlier
than 0.3s (yellow region), where the 
right-of-way is ambiguous. Ego is likely to stop if the other agent arrives earlier
(red region). If the other
agent arrives earlier by more than 1.5s (green region), the intersection is clear
for the ego
and the likelihood of proceeding forward goes back up. This experiment shows
that in ambiguous situations, ego attempts to drive forward, just like a human
driver would in the absence of right-of-way rules. At the same time, ego does
stop if a collision is more likely.

Next we study ego's behavior in safety-critical situations. 
\cref{fig:midas_vanilla_min_ttc3_ego_go_pcts} compares the percentage
of timesteps where ego goes forward (Y-axis) against the minimum TTC
across all agents in the vicinity
if both ego and the other agent decide to go forward (X-axis).
For small TTC when collision is immediate, MIDAS is less likely to
proceed than MLP. This shows that
MIDAS does not blindly drive forward in safety critical situations.

\begin{figure}[!h]%
\centering
\captionsetup[subfigure]{justification=centering}
\begin{subfigure}[t]{.17\textwidth}
    \includegraphics[width=\linewidth]{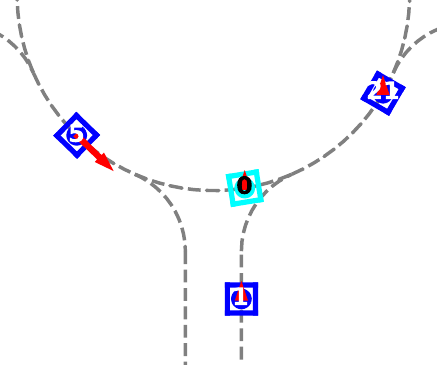}
\caption{Car-following episode scene}
\label{fig:sim_actions_basic_927_addon_render}
\end{subfigure}
\hspace{1em}
\begin{subfigure}[t]{.24\textwidth}
    \includegraphics[width=\linewidth]{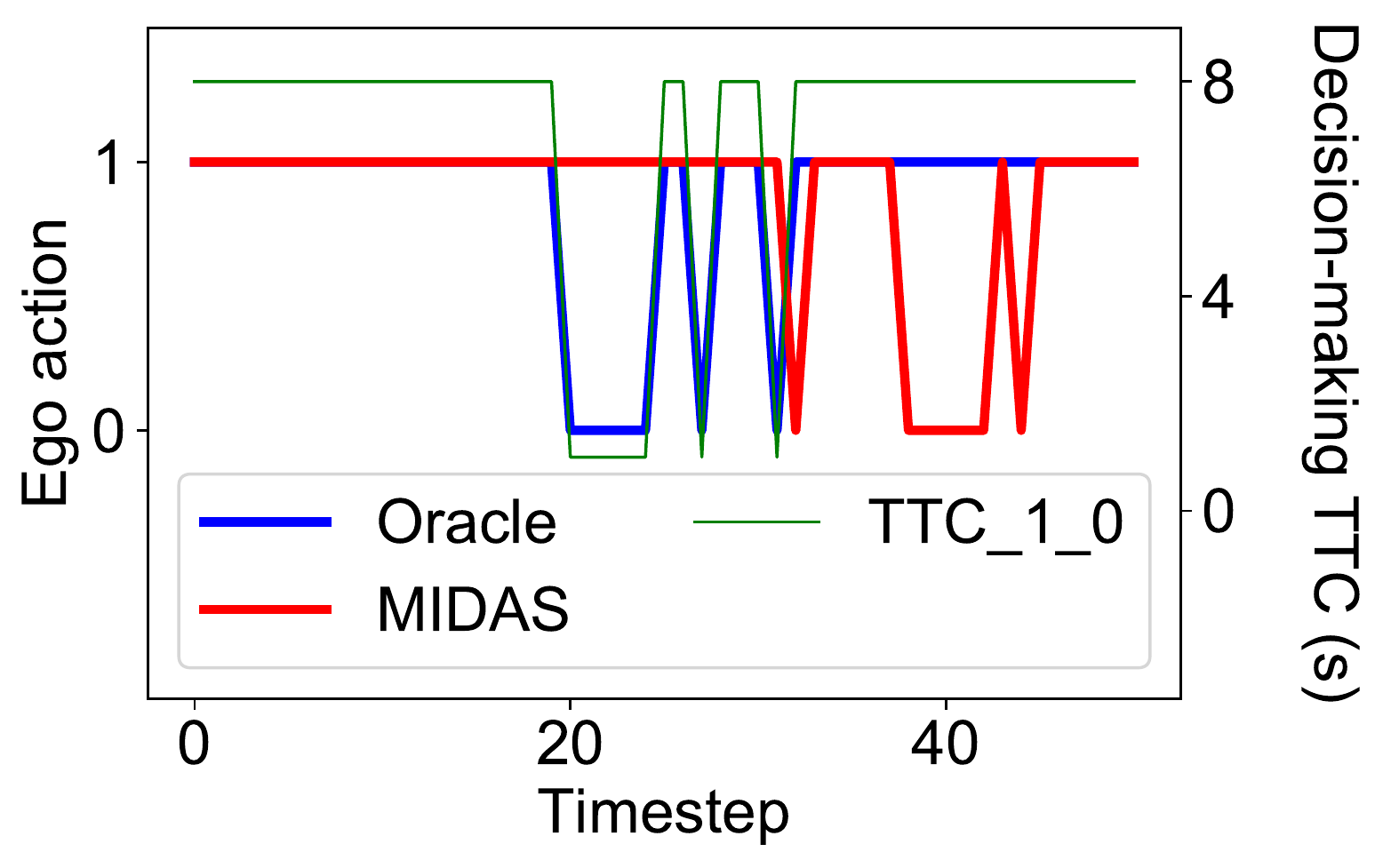}
\caption{Car-following actions and decision-making TTC}
\label{fig:sim_actions_basic_927_addon_plot}
\end{subfigure}
\hspace{1em}
\begin{subfigure}[t]{.18\textwidth}
    \includegraphics[width=\linewidth]{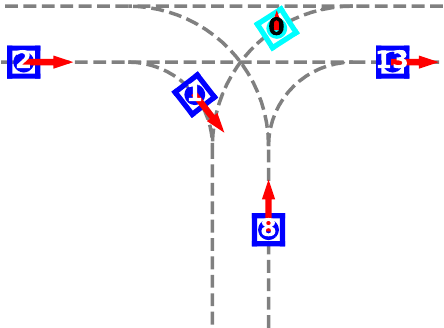}
\caption{Left-turn episode scene}
\label{fig:sim_actions_basic_873_addon_render}
\end{subfigure}
\hspace{1em}
\begin{subfigure}[t]{.24\textwidth}
    \includegraphics[width=\linewidth]{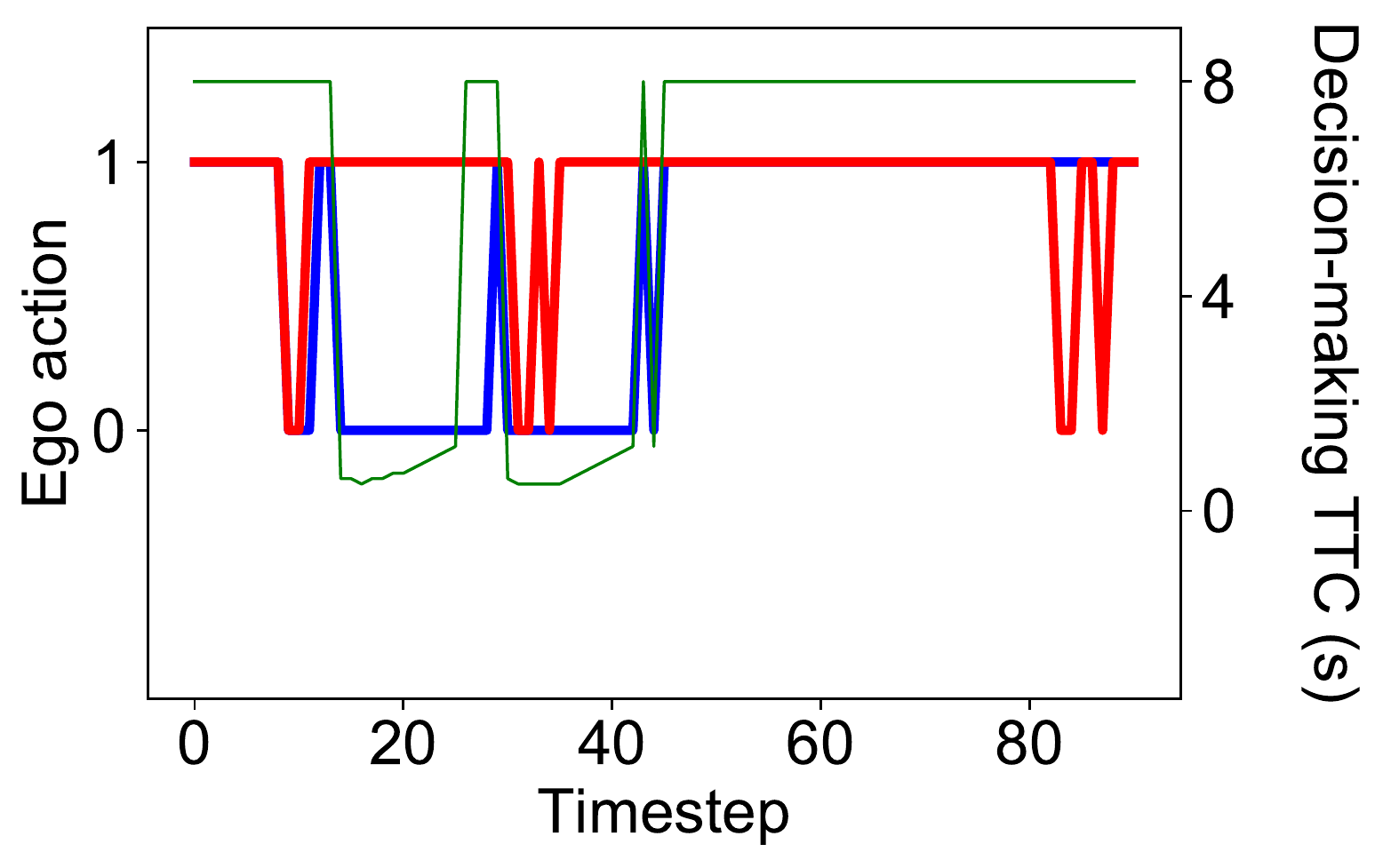}
\caption{Left-turn actions and decision-making TTC}
\label{fig:sim_actions_basic_873_addon_plot}
\end{subfigure}
\caption{\tbf{Typical episodes where MIDAS differs from Oracle.} Ego is cyan; all other agents are blue. In car-following, Oracle stops for front vehicle at a far distance while blocking traffic at an intersection, but MIDAS chooses to stop later. In left-turn, Oracle waits until there's a big clearance after agent 1 turns right before proceeding, but MIDAS drives forward right after. Both episodes illustrate how MIDAS drives more efficiently. Plots show Oracle (blue) actions, MIDAS (red) actions and $\tau^{1j}_{1,0}$ (green).}
\label{fig:results_sim_action_plots}%
\end{figure}

\textbf{4. Does the model make decisions properly in highly interactive situations?}
We define highly interactive situations for agent $i$ as those where $\tau^{ij}_{1,0}$
(green line in~\cref{fig:sim_actions_basic_927_addon_plot},
~\cref{fig:sim_actions_basic_873_addon_plot}) is lower than a threshold,
indicating that there's at least one other agent very close
by which forces $i$ to stop. We run Oracle on the test set and \emph{record
what the learned model would have done if it were in the same situation} as the Oracle.
In car-following (\cref{fig:sim_actions_basic_927_addon_render}), Oracle stops for the front vehicle a long 
distance away while blocking the traffic at an intersection, but MIDAS chooses to move forward and stop closer 
to the front vehicle. In left-turn (\cref{fig:sim_actions_basic_873_addon_render}), Oracle waits until there's a big clearance after agent 1 turns right before proceeding, but MIDAS drives forward right after. Both episodes 
show how MIDAS drives more efficiently. Refer to~\cref{app:typical_episodes} for
illustrations.

\begin{figure*}[!htpb]
\centering
\begin{subfigure}[b]{.22\textwidth}
    \includegraphics[width=\linewidth]{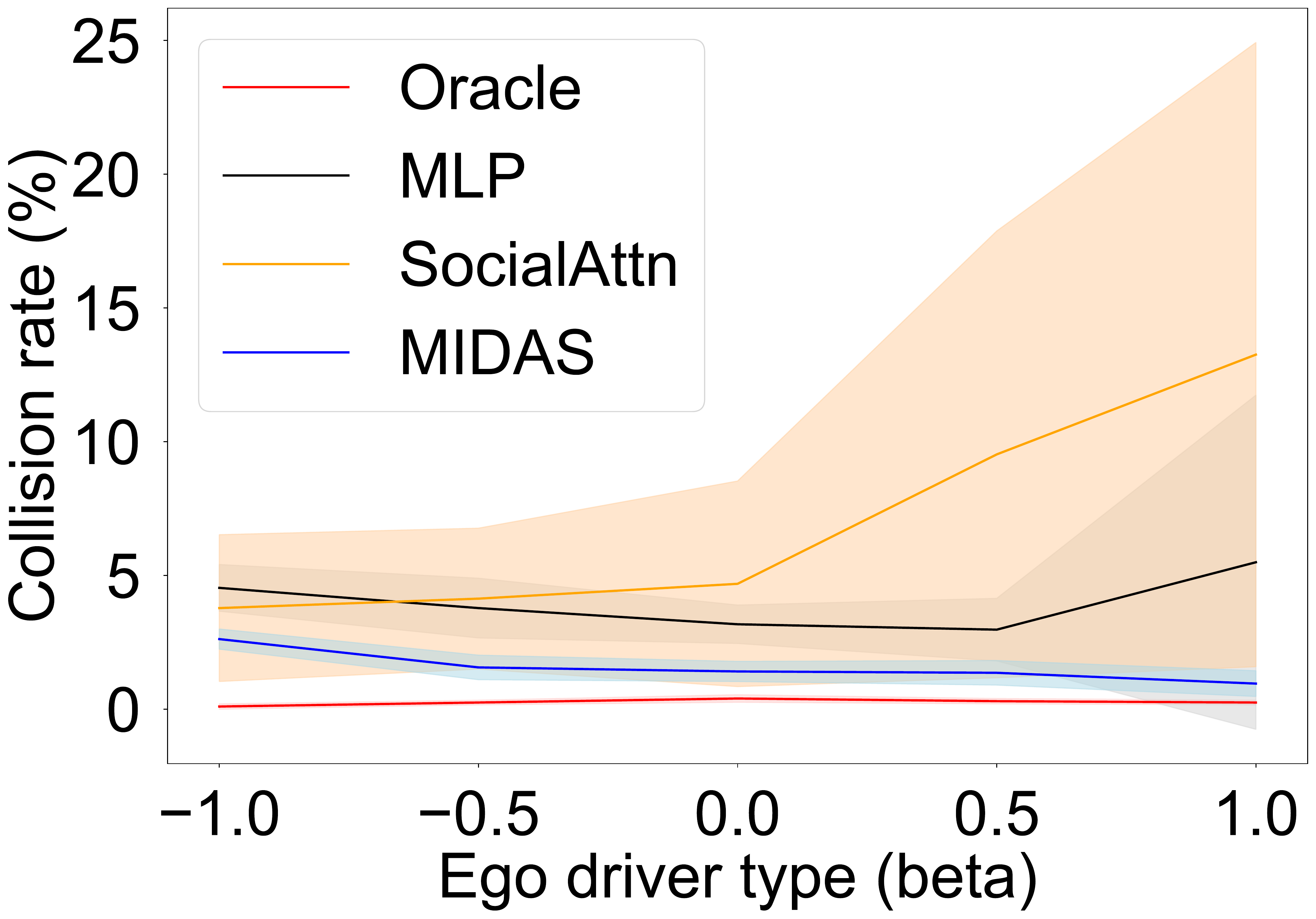}
\caption{}
\label{fig:fixed_ego_b1_tr5_eval_set_cls}
\end{subfigure}
\begin{subfigure}[b]{.22\textwidth}
    \includegraphics[width=\linewidth]{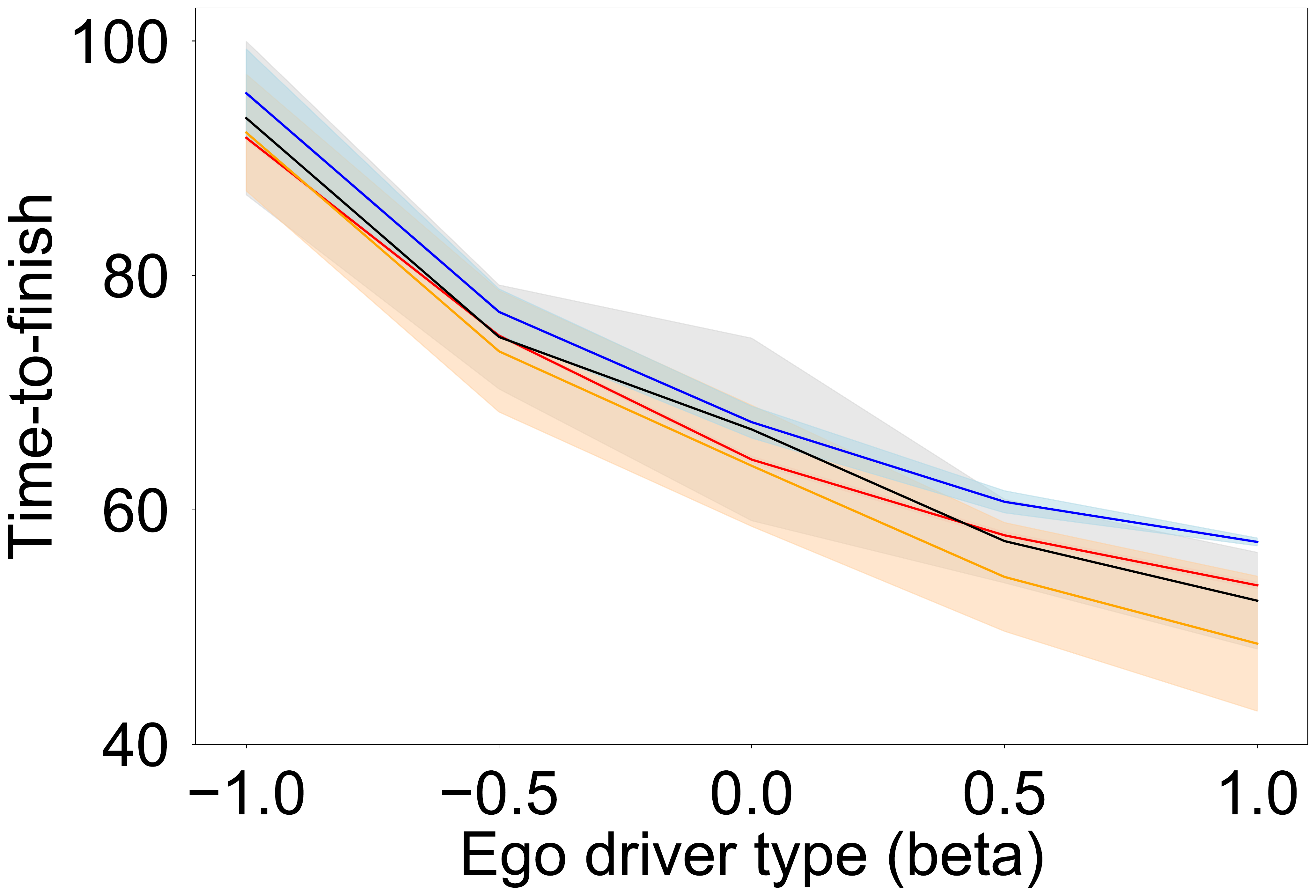}
\caption{}
\label{fig:fixed_ego_b1_tr5_eval_set_ttf}
\end{subfigure}
\begin{subfigure}[b]{.53\textwidth}
    \begin{table}[H]
    \renewcommand{\arraystretch}{1.2}
    \resizebox{\textwidth}{!}{
        \begin{tabular}{|l|p{2cm}|r|r|r|}
        \toprule
        \textbf{Model} & \textbf{Intersection} &
        \textbf{Collision (\%)} & \textbf{Timeout (\%)} & \textbf{Success (\%)} \\
        \midrule
        \multirow{2}{*}{Car-Follower}  & T-intersection    & 4.20 $\pm$ 0.62 & 0.00 $\pm$ 0.00 & 95.80 $\pm$ 0.62 \\
                                 & Roundabout        & 2.88 $\pm$ 0.38 & 0.00 $\pm$ 0.00 & 97.12 $\pm$ 0.38 \\
        \hline
        \multirow{2}{*}{MLP}  & T-intersection    & 2.66 $\pm$ 1.00 & 2.01 $\pm$ 1.89 & 95.33 $\pm$ 2.71 \\
                                 & Roundabout        & 3.38 $\pm$ 2.33 & 0.00 $\pm$ 0.00 & 96.62 $\pm$ 2.33 \\
        \hline
        \multirow{2}{*}{MIDAS}  & T-intersection    & 1.23 $\pm$ 0.59 & 0.39 $\pm$ 0.13 & 98.38 $\pm$ 0.62 \\
                                 & Roundabout        & 1.35 $\pm$ 1.01 & 0.68 $\pm$ 0.75 & 97.97 $\pm$ 0.39 \\
        \bottomrule
        \end{tabular}
    }
    \caption{}
    \label{tab:results_intersection_type_pfmc}
    \end{table}
    \end{subfigure}
\caption{\tbf{MIDAS' strategy is adaptive.} \cref{fig:fixed_ego_b1_tr5_eval_set_cls,fig:fixed_ego_b1_tr5_eval_set_ttf} shows that MIDAS plans more efficiently and is consistently safer than all other learned models as ego driver type changes. Deep Set and Car-Follower both perform worse than MIDAS and are not shown here for clarity. SocialAttn refers to SocialAttention. \cref{tab:results_intersection_type_pfmc} shows test performance of Car-Follower, MLP and MIDAS at different intersections.}%
\label{fig:results_fixed_ego_b1}%
\end{figure*}

\tbf{5. Are MIDAS' strategies adaptive? How does model performance change with driver-type?}
At test time, we change the ego driver type to $-1,-0.5,0,0.5,1$ and run it on
the test set with all other agents running the Oracle planner. The results are
shown in \cref{fig:results_fixed_ego_b1}. MIDAS plans more efficiently as ego
driver type increases, as shown in the decreasing time-to-finish (\cref{fig:fixed_ego_b1_tr5_eval_set_ttf}), while maintaining the collision rate
almost constant (\cref{fig:fixed_ego_b1_tr5_eval_set_cls}). \cref{fig:fixed_ego_b1_tr5_eval_set_cls} also shows that MIDAS is consistently safer than all other learned models across different ego driver types. MIDAS timeout rate is relatively constant across ego driver types and is shown in \cref{app:timeout_rate_across_driver_types}.

\tbf{6. MIDAS performs consistently across different driving scenes.}
\cref{tab:results_intersection_type_pfmc} shows that the collision
rate of MIDAS is consistently lower than other models.
The collision/timeout rates also show only a small variation across
intersection types.

\section{Related Work}
\label{s:related_work}

Intention-aware planning with known dynamics for agents
can be formulated as a
POMDP, see e.g.,~\cite{bandyopadhyay2013intention} and solved using point-based solvers~\cite{bai2015intention}. The focus is typically on
planning a safe trajectory and not on interaction~\cite{wongpiromsarn2012incremental}.
For urban driving, one may assume knowledge of
road safety rules and use game-theoretic formulations to model multi-agent
decision making~\cite{chaudhari2014incremental,zhu2014game}.
Interactive, model-free highway driving using deep RL is popular, e.g.,
policies for dense traffic using a
prediction model~\cite{henaff2019model}, or ones that incorporate actions of
other agents~\cite{rhinehart2019precog}. These papers involve fewer and easier
interactions than ours.
%
Merging has been studied in~\cite{schmerling2018multimodal,baecooperation}
using predictive models for model predictive control, estimating future horizon cost via exhaustive search or Monte-Carlo rollout.
MIDAS is complementary to these approaches and although we do not use a prediction model
to narrow down our focus to interactions, such a model can be easily incorporated
using model-based RL techniques.

Our use of adjustable driver type is similar to~\cite{hu2019interaction}
where aggressive agents are
encouraged to merge faster by overtaking others. This work however uses self-play
to train the policy and while this approach is
reasonable for highway merging, the competition is likely to result in
high collision rates in busy urban
intersections such as ours.

\cite{Sadigh16} uses inverse RL to learn the cost function of
human drivers and uses it to influence other drivers. This method
is computationally expensive and is limited to simplistic interactions.
One may improve this by separating long-term and near-time planning
at the cost of optimality~\cite{fisac2019hierarchical}.
In comparison, our attention-based model
can scale to interaction with a large number of agents;
our quantitative evaluation methods are also
more thorough.

Learning-based approaches use featurization to tackle different
lane geometries~\cite{li2019urban}, roundabouts~\cite{chen2019model},
or use simplified sequential action representations~\cite{isele2018navigating}.
The observation vector is a concatenation of the states of
other agents~\cite{schmerling2018multimodal}, spatial pooling
that aggregates past trajectories of each agent~\cite{gupta2018social},
or birds-eye-view rasterization~\cite{tang2019towards, bansal2018chauffeurnet}.
In contrast, the set-transformer in MIDAS is an easy, automatic
way to encode variable-sized observation vectors.
In this sense, our work is closest to ``Social Attention'' in~\cite{leurent2019social} which learns to influence other agents
based on road priority and demonstrates results on a limited set of road geometries;
MIDAS compares favorably to this method in~\cref{sec:experiments}.

\section{Discussion}

We studied interactions of autonomous vehicles with other vehicles
in urban driving scenarios. We used deep RL 
techniques to learn a policy for an autonomous agent
that can influence the actions of other agents.
Interaction-aware autonomous driving is a
pertinent problem but it is difficult to analyze systematically
across different scenarios.
Our work is a step towards improving the status quo.
Its key features are
user-tunable adaptive policies that are
trained efficiently, the ability to pay attention to only
the part of the observation vector that matters for control irrespective
of the number of other agents in the vicinity,
and an
evaluation methodology that can answer difficult questions like
``does the ego agent influence other agents'', or ``how does the agent perform
in highly interactive settings''
in a quantitative and systematic manner.


Our goal is to translate interaction-aware planning from simulation to reality.
On the algorithmic side, the main challenge is to build intent-aware
prediction models and perform probabilistic reasoning over their
outcomes while building the policy. As far as an implementation
on autonomous vehicles is concerned,
we are cognizant of the fact that deep RL policies have a very high variance in 
their real-world performance.
Indeed, any learning-based approach is essentially blind to behaviors \emph{not}
present in the data.
Our work should be thought of as providing
a prior---that can be tuned via the driver-type to be more optimistic than a
worst-case assumption---for
existing planning algorithms for autonomous driving that build upon
rapidly-exploring random trees~\cite{kuwata2008motion,censi2019liability}, model predictive control~\cite{urmson2008autonomous} etc.

{
\clearpage
\small
\setlength{\bibsep}{0.5em}
\bibliography{references}  
}

\clearpage
\appendix
\begin{center}
    \Large \tbf{Supplementary Material\\
        \mytitle}
\vspace*{2em}
\end{center}

\section{Approach}
\subsection{Attention-based Policy Architecture}
\begin{figure}[h!]%
    \centering
    \captionsetup[subfigure]{justification=centering}
    \includegraphics[width=0.5\linewidth]{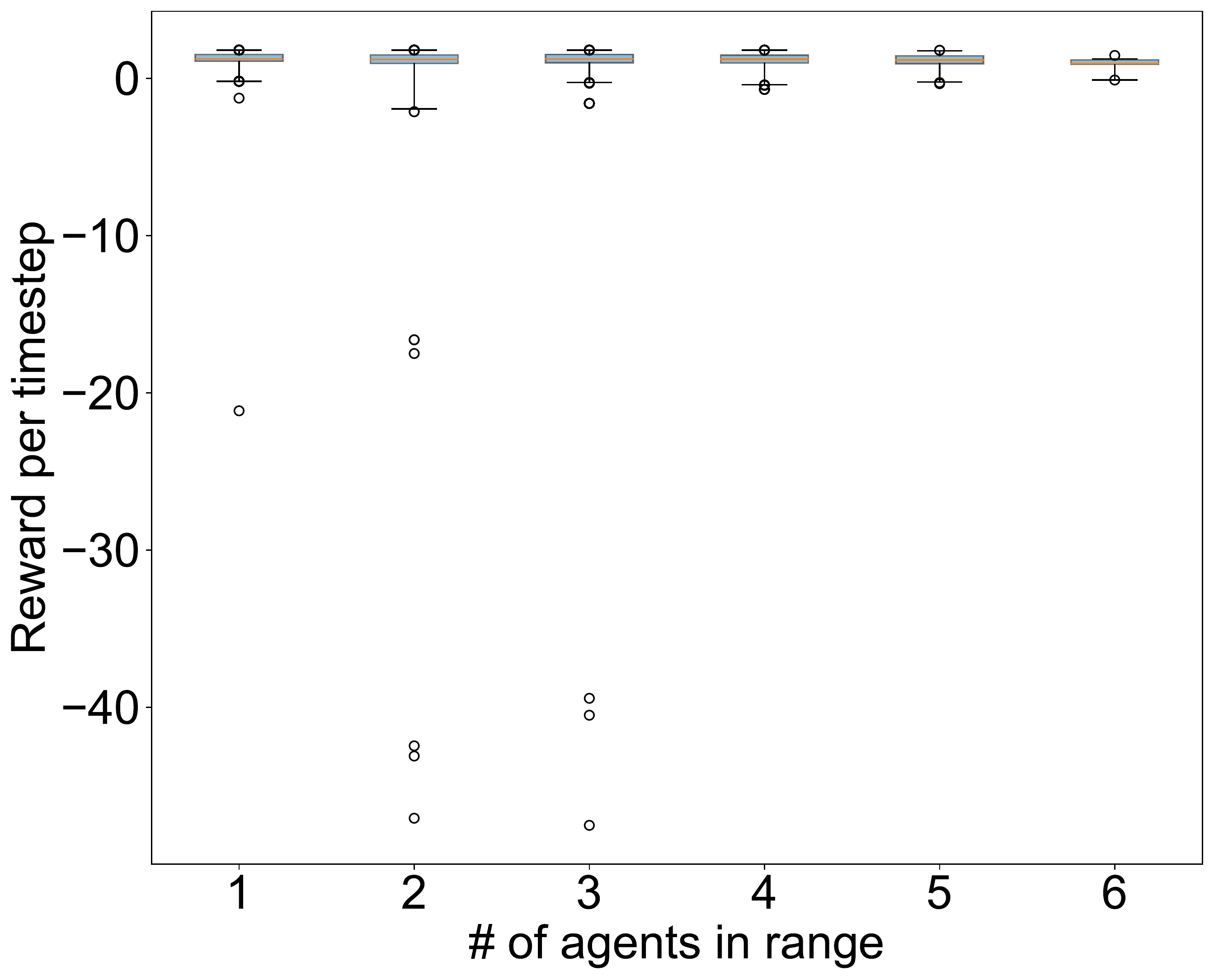}
    \caption{\tbf{MIDAS performance is consistent across situations with different difficulty.} Box plot of the reward \emph{received} per timestep against agent density in ego's vicinity. The box plot whisks represent 0.5\% and 99.5\% percentiles.}
    \label{fig:reward_density_plot}
\end{figure}

It's important for the ego to be able to handle a varying number of other agents. We run one seed of MIDAS on the test set and plot the reward received per timestep against the agent density 
in ego's vicinity, represented by the number of agents in ego's 
observation range. Higher density indicates higher difficulty, 
because ego needs to pay attention to the more important agents. As shown in~\cref{fig:reward_density_plot}, MIDAS performance is
consistent across situations with different difficulty. This 
suggests that the attention mechanism works well; it is able to 
pay attention to only relevant parts of the observation vector 
even when there is a large number of agents in the vicinity.

\subsection{Designing the Reward Function}
\label{app:reward_coefficients}
The weights and biases of the reward function described in \cref{ss:reward_design} are listed in \cref{eq:reward}. The adaptive cruise control penalty penalizes the agent for following too closely to the agent in front and is designed such that it's capped between -2 and 0. 
The reward coefficients of all sub-rewards, except for stalemate penalty, are chosen using a generic RL agent, which has the same structure as MLP but does not use two copies of parameters. The agent was trained using the standard time-lagged Q-learning algorithm without the two variants introduced in~\cref{ss:off_policy}. We added the stalemate penalty later to reduce the occurrence of timeouts across all models that were being trained.

\beq{
    \re(x_t; \xe_g, \be) =
    \begin{cases}
    -0.05 \be - 0.15    & \trm{time penalty for every every timestep}\\
    0.5 \be + 1.5       & \trm{if ego speed is non-zero}\\
    -5 \be - 20         & \trm{timeout penalty if}\ t = T_{\trm{max}}\\
    -0.5 \be - 1.5      & \trm{stalemate penalty}\\
    -5 \be - 45         & \trm{collision penalty}\\
    -2 + \f{2}{1 + e^{-d_{\trm{follow}}}} & \trm{if ego within distance}\ \delta_{\trm{follow}}\ \trm{of the car ahead of it}
    \end{cases}
    \label{eq:reward}
}


\clearpage
\section{Evaluation Methodology}
\subsection{Types of Interaction Episodes}
\begin{figure}[h!]%
    \centering
    \captionsetup[subfigure]{justification=centering}
    \begin{subfigure}[t]{.24\textwidth}
        \includegraphics[width=\linewidth]{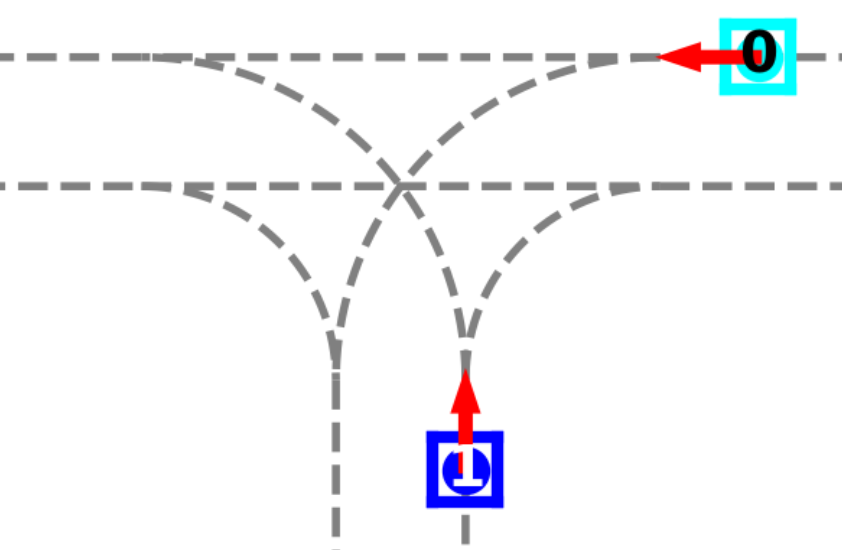}
        \caption{Setting 1}
        \label{}
    \end{subfigure}
    \begin{subfigure}[t]{.3\textwidth}
        \includegraphics[width=\linewidth]{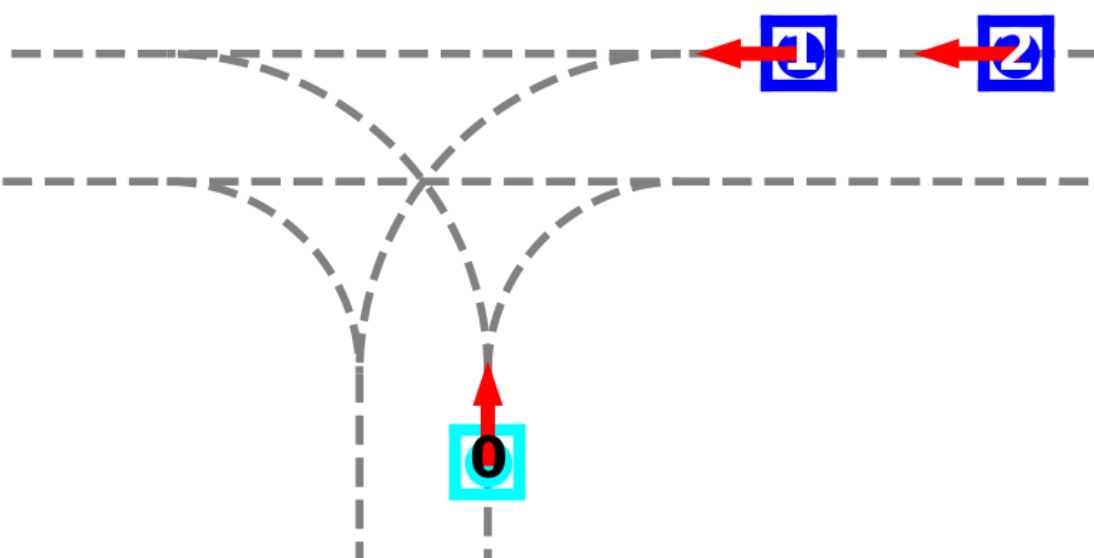}
        \caption{Setting 2}
        \label{}
    \end{subfigure}
    \begin{subfigure}[t]{.23\textwidth}
        \includegraphics[width=\linewidth]{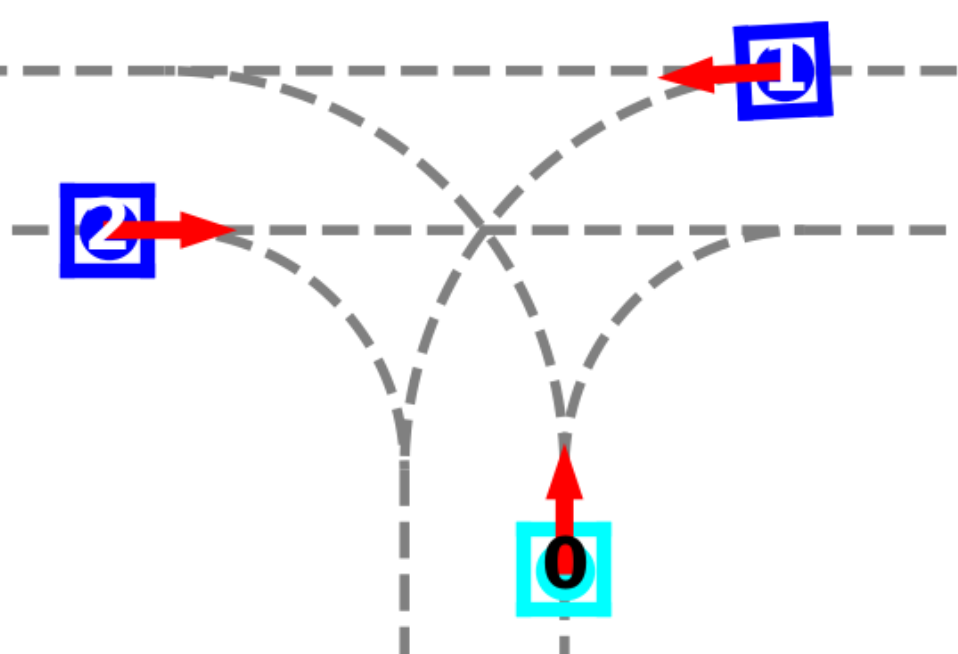}
        \caption{Setting 3}
        \label{}
    \end{subfigure}
    \caption{The interaction settings used for generating the interaction set. In Setting 1 and 2, agent 0 and 1 will simultaneously arrive at the lane intersection. In Setting 2, agent 2 is at the minimum following distance behind agent 1. In Setting 3, all three agents will arrive simultaneously at the lane intersection.}
    \label{fig:3-3_interaction_settings}
\end{figure}

~\cref{fig:3-3_interaction_settings} shows the three basic types of interaction episodes described in~\cref{ss:evaluation}.
\label{app:interaction_settings}

\subsection{Training, Validation and Test Set Configurations}
\label{app:train_validation_test_configs}
We use a mix of 25\% generic episodes, 25\% collision episodes, and 50\%
interaction episodes for training and a validation set composed of 100 random
and 100 interaction episodes. For
reporting the test performance,
we use 250 generic episodes and 250 interaction episodes.
This evaluation
benchmark reflects general driving
scenarios. We also use the test interaction set, which contains 381 interaction episodes, to evaluate model performance on interactive scenarios.

\clearpage
\section{Model Implementation Details}
\label{app:model_implementation_details}

\subsection{Model Architecture}
\label{app:appendix_midas}

\begin{figure}[h!]%
    \centering
    \includegraphics[width=0.8\linewidth]{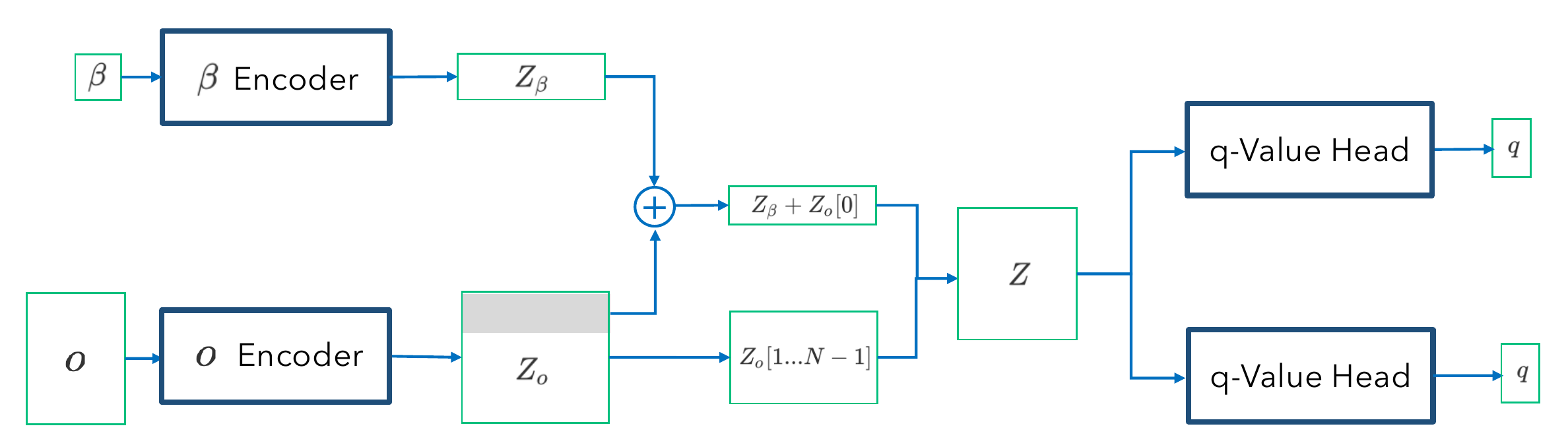}
    \caption{The architecture of MLP, DeepSet, SocialAttention and MIDAS. $\be, o$ refers to ego driver type and observation, respectively. The two values are encoded separately, and then the encoded $\be$ is added to the first row of the encoded $o$. After that, the encoding is passed into two q-value heads with identical structures, each outputting an estimate of the q-values of the two actions, mentioned in~\cref{ss:off_policy}.}
    \label{fig:model_architecture}
\end{figure}

The model architecture is shown in~\cref{fig:model_architecture}, and the detailed implementation of the models experimented with in~\cref{sec:experiments} are shown in~\cref{tab:model_implementation}.

\begin{table}[H]
\centering
    \renewcommand{\arraystretch}{1.0}
    \caption{Implementation of the components in the model shown in~\cref{fig:model_architecture}. The values in [] refer to input and output dimensions of fully-connected layers. $|o|$ refers to the dimension of the total state vector with all agent information, while $|o^k|$ refers to the dimension of the state vector of a single agent. ``Int. Layer'' refers to ``Intermediate Layer'', applied to the combined encoding $Z$ in~\cref{fig:model_architecture} before it's passed into the two q-value heads.
    SocialAttention modules: Based on~\cite{leurent2019social}, the parameters in EgoAttention correspond to: hidden dimension, number of heads.
    MIDAS modules: Based on~\cite{lee2018set}, the parameters in ISAB correspond to: input dimension, output dimension, number of heads, number of induced vectors, whether to apply LayerNorm (T/F); the parameters in PMA correspond to: input dimension, number of heads, number of seeds, whether to apply LayerNorm (T/F); the parameters in SAB correspond to: input dimension, output dimension, number of heads, whether to apply LayerNorm (T/F).}
    \label{tab:model_implementation}
    \vspace*{1em}
    \resizebox{0.9 \textwidth}{!}{
        \small
        \begin{tabular}{|p{2.3cm} l l l l|}
        \toprule
        & \textbf{MLP} & \textbf{DeepSet} & \textbf{SocialAttention} & \textbf{MIDAS}\\
        \midrule
            $o$ Encoder
          & [$|o|$,128] & [$|o^k|$,128]& [$|o^k|$,64] & ISAB($|o^k|$, 128, 4, 32, T)\\
          & ReLU      & ReLU     & ReLU & ISAB(128, 128, 4, 32, T)\\
          & [128,128] & [64,64] & [64,64] & \\
          &           & ReLU    & ReLU & \\
          &           & [64,128] & [64,64] & \\
         \hline
         $\beta$ Encoder
          & [1,64]  & [1,64]  & [1,64] & [1,64]\\
          & ReLU    & ReLU    & ReLU & ReLU \\
          & [64,128]& [64,128]& [64,64] & [64,128]\\
          & & & ReLU & ReLU \\
          & & & [64,64] & [128,128]\\
         \hline
         Int. Layer
          & / & [128,128] & [64,64] & / \\
          &   & ReLU      & ReLU &  \\
          &   & [128,128] & [64,64] &\\
          &   &  & EgoAttention(64,2) & \\
         \hline
         $q$-Value Head
          & [128,128] & [128,128] & [64,64] & PMA(128, 4, 2, T) \\
          & ReLU & ReLU & ReLU & SAB(128, 128, 4, T) \\
          & [128,2] & [128,2] & [64,2] & SAB(128, 128, 4, T) \\
          &  &  &  & [128,1]\\
        \bottomrule
        \end{tabular}
    }
    \end{table}

\subsection{Hyper-parameters}
\label{app:hyperparameters}
Hyper-parameters used for training: $\gamma=0.99$; policy networks are updated at the end of every episode for the same number of time steps as the episode; the time-lagged network $\theta_{\text{lag}}$ is updated every $100$ training steps using $\tau=0.2$; replay buffer size $= 200000$; batch size $= 128$; Adam optimizer is used with learning rate $= 2e-5$; $\epsilon$-greedy is used for the first $500$ time steps, where $\epsilon$ is annealed exponentially from $1.0$ to $0.01$. To ensure fair comparison, the hyper-parameters are the same across all four models: MLP, Deep Set, Social Attention, MIDAS.

\clearpage
\section{Results}
\subsection{Training Curves}
\label{app:training_curves}
\begin{figure}[h!]%
\centering
\captionsetup[subfigure]{justification=centering}
\begin{subfigure}[t]{.3\textwidth}
    \includegraphics[width=\linewidth]{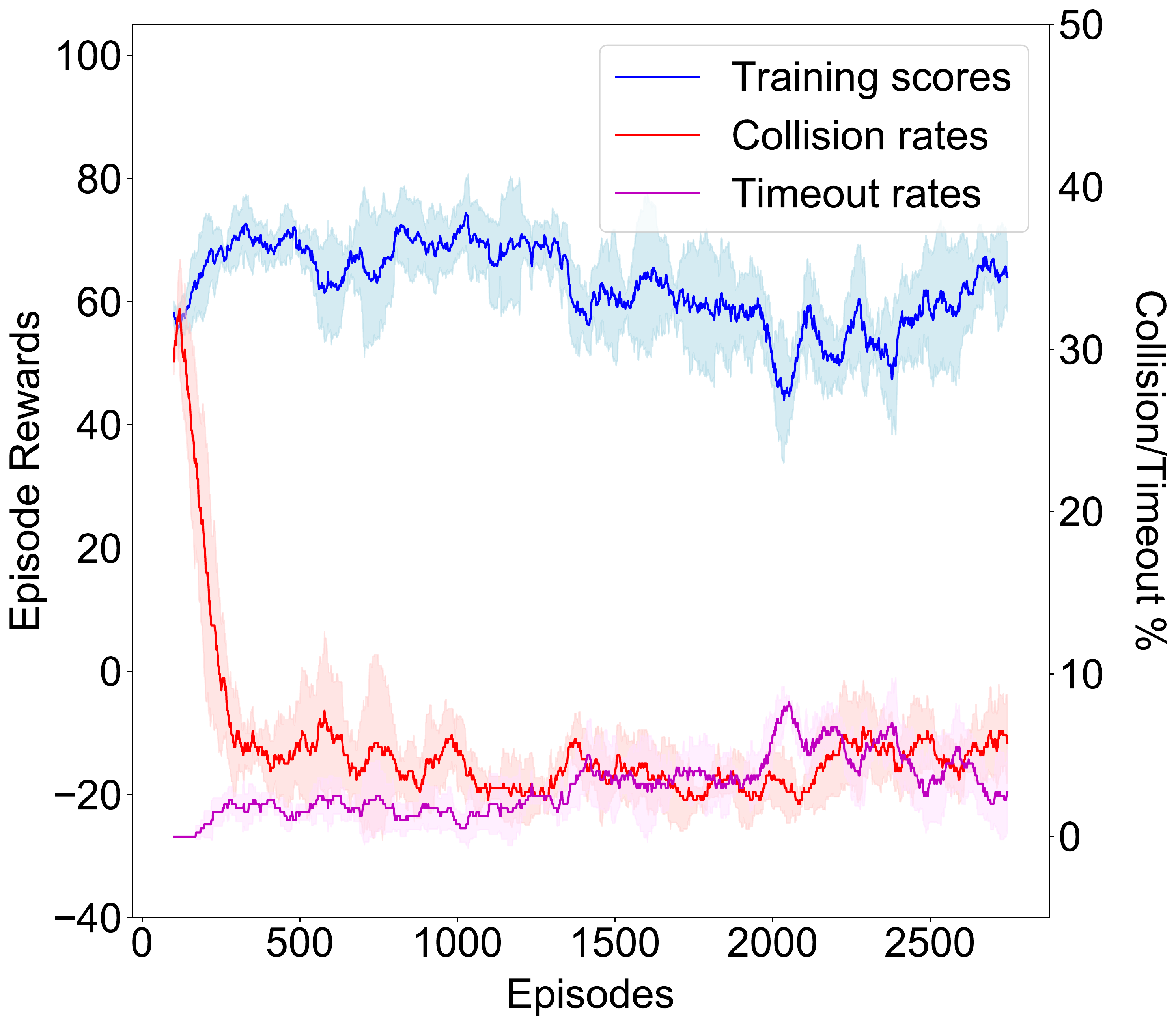}
\caption{MLP}
\label{}
\end{subfigure}
\begin{subfigure}[t]{.3\textwidth}
    \includegraphics[width=\linewidth]{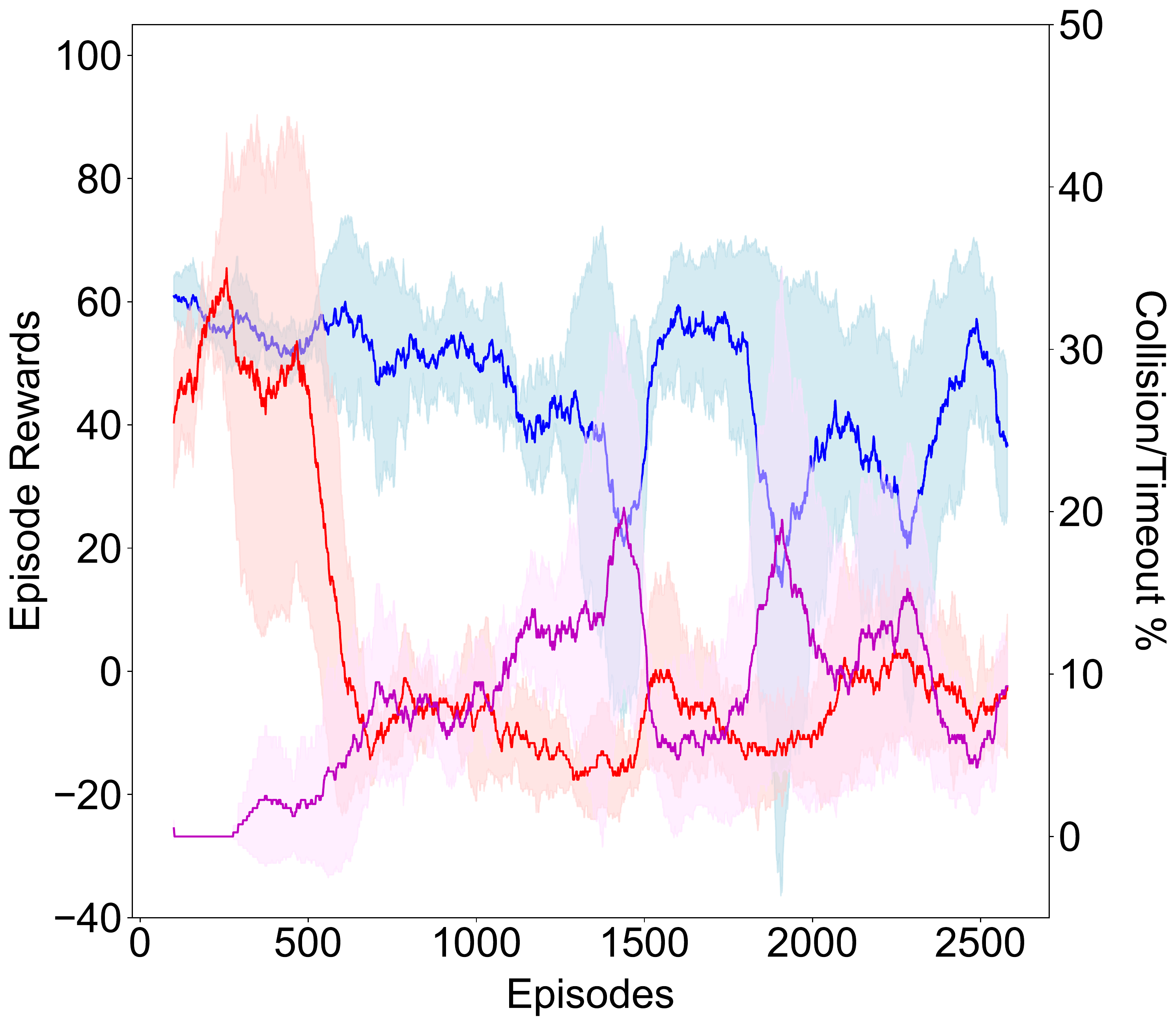}
\caption{DeepSet}
\label{}
\end{subfigure}

\begin{subfigure}[t]{.3\textwidth}
    \includegraphics[width=\linewidth]{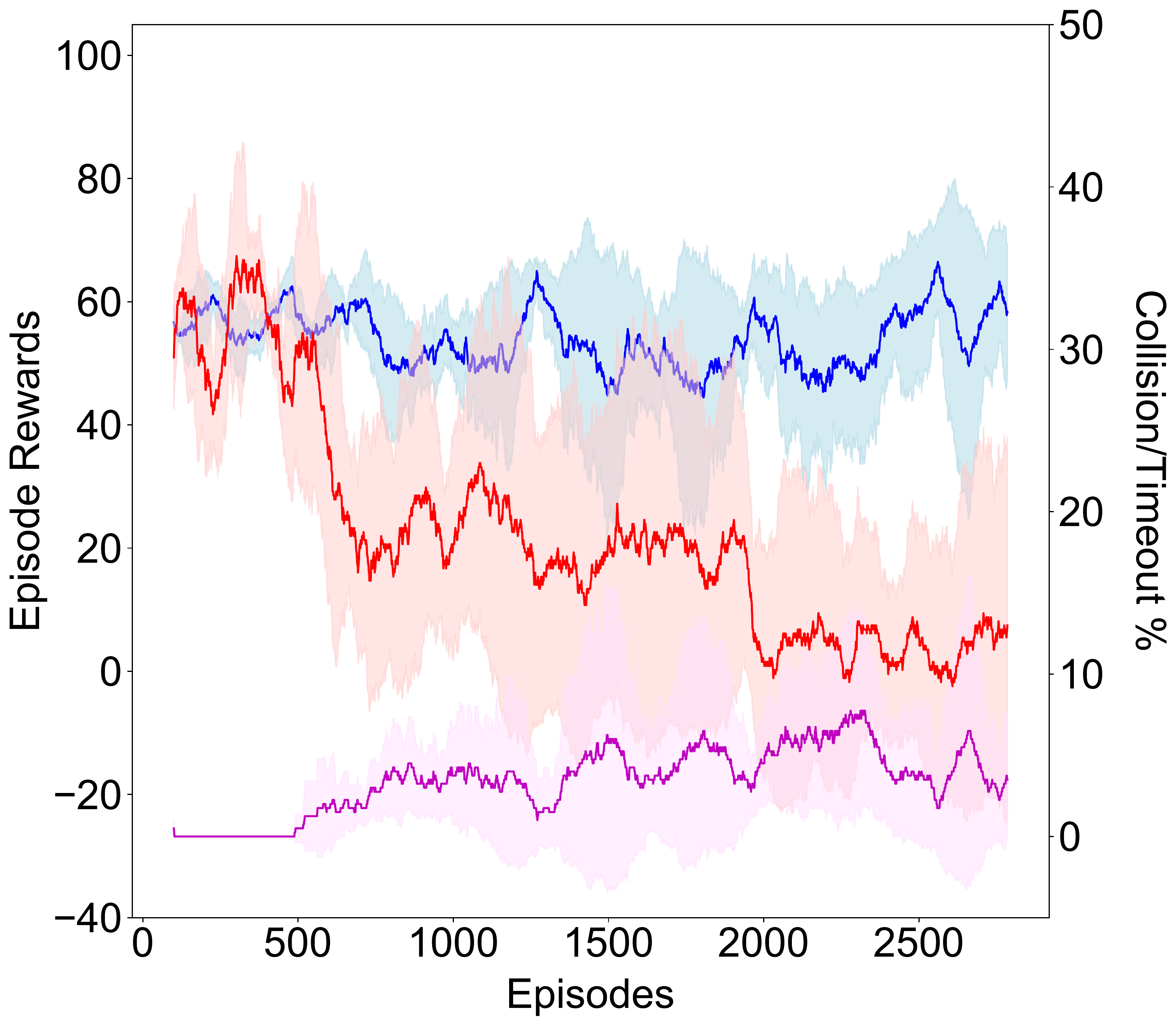}
\caption{SocialAttention}
\label{}
\end{subfigure}
\begin{subfigure}[t]{.3\textwidth}
    \includegraphics[width=\linewidth]{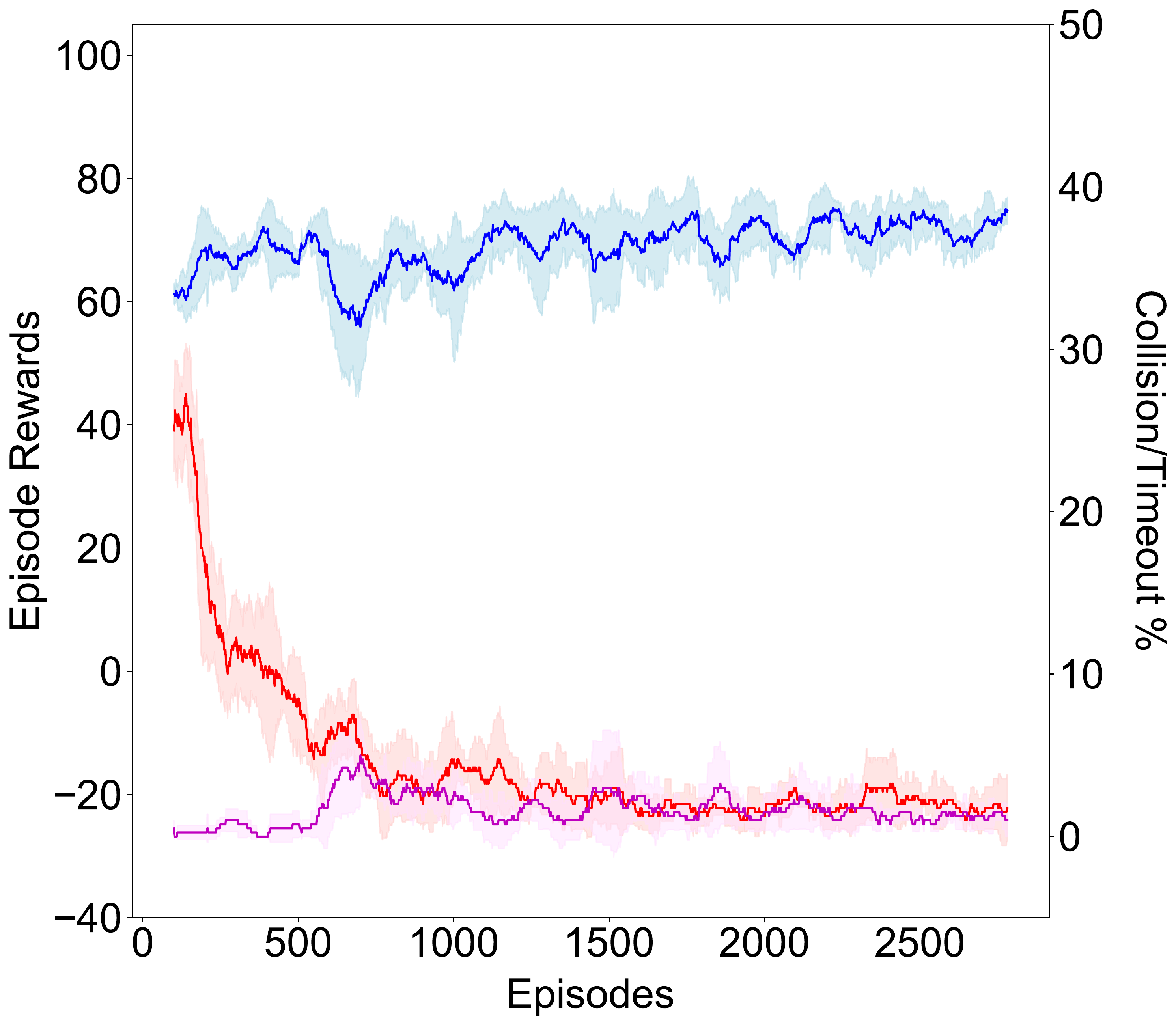}
\caption{MIDAS}
\label{}
\end{subfigure}
\caption{Training curves of the models with different attention mechanisms.}
\label{fig:6-1_training_curves}%
\end{figure}

\cref{fig:6-1_training_curves} shows the training curves of MLP, DeepSet, SocialAttention and MIDAS.

\clearpage
\subsection{Typical Simulated Episodes}
\begin{figure}[h!]%
\centering
\captionsetup[subfigure]{justification=centering}
\begin{subfigure}[t]{.3\textwidth}
    \includegraphics[width=\linewidth]{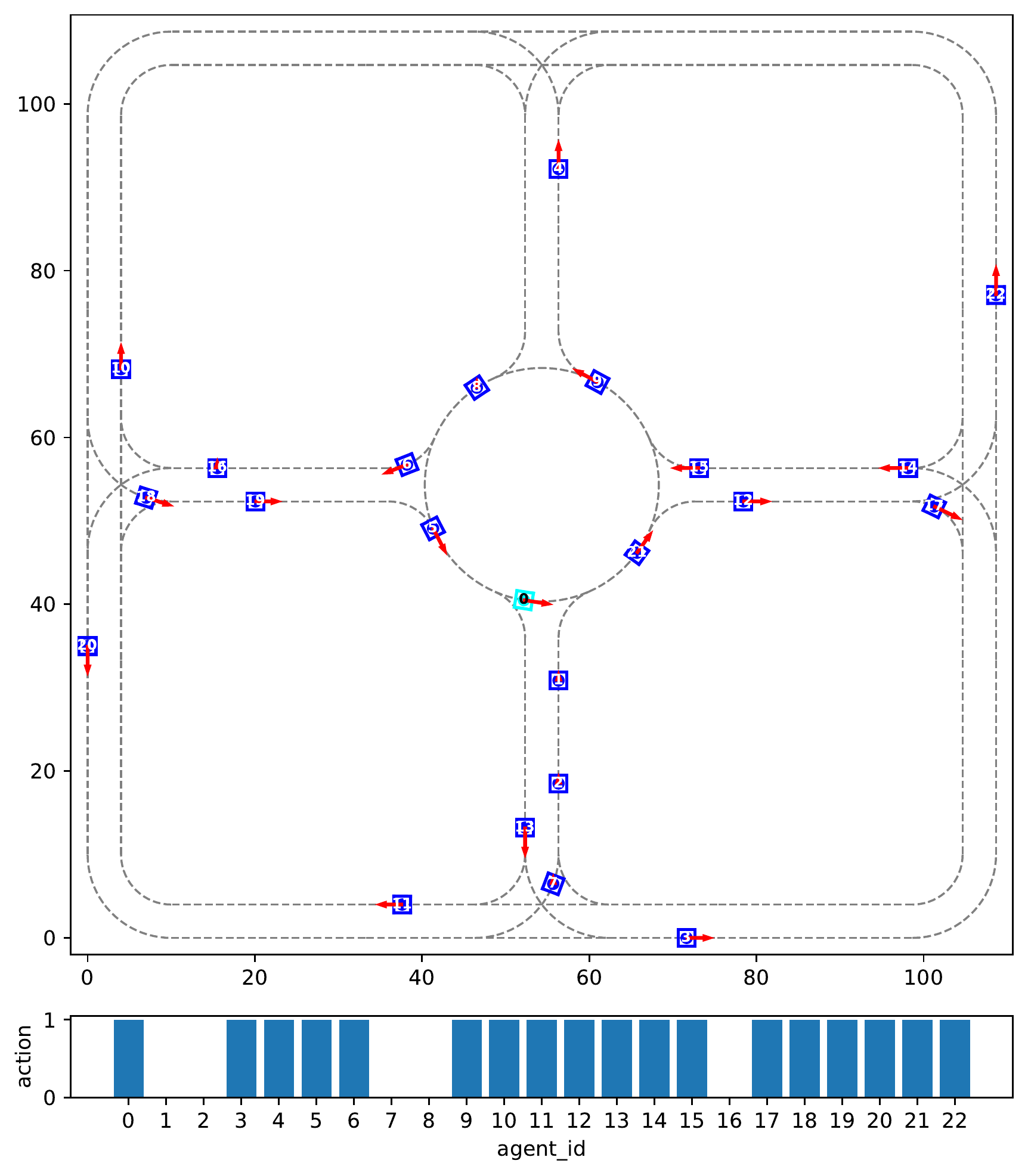}
\caption{t=15}
\label{}
\end{subfigure}
\begin{subfigure}[t]{.3\textwidth}
    \includegraphics[width=\linewidth]{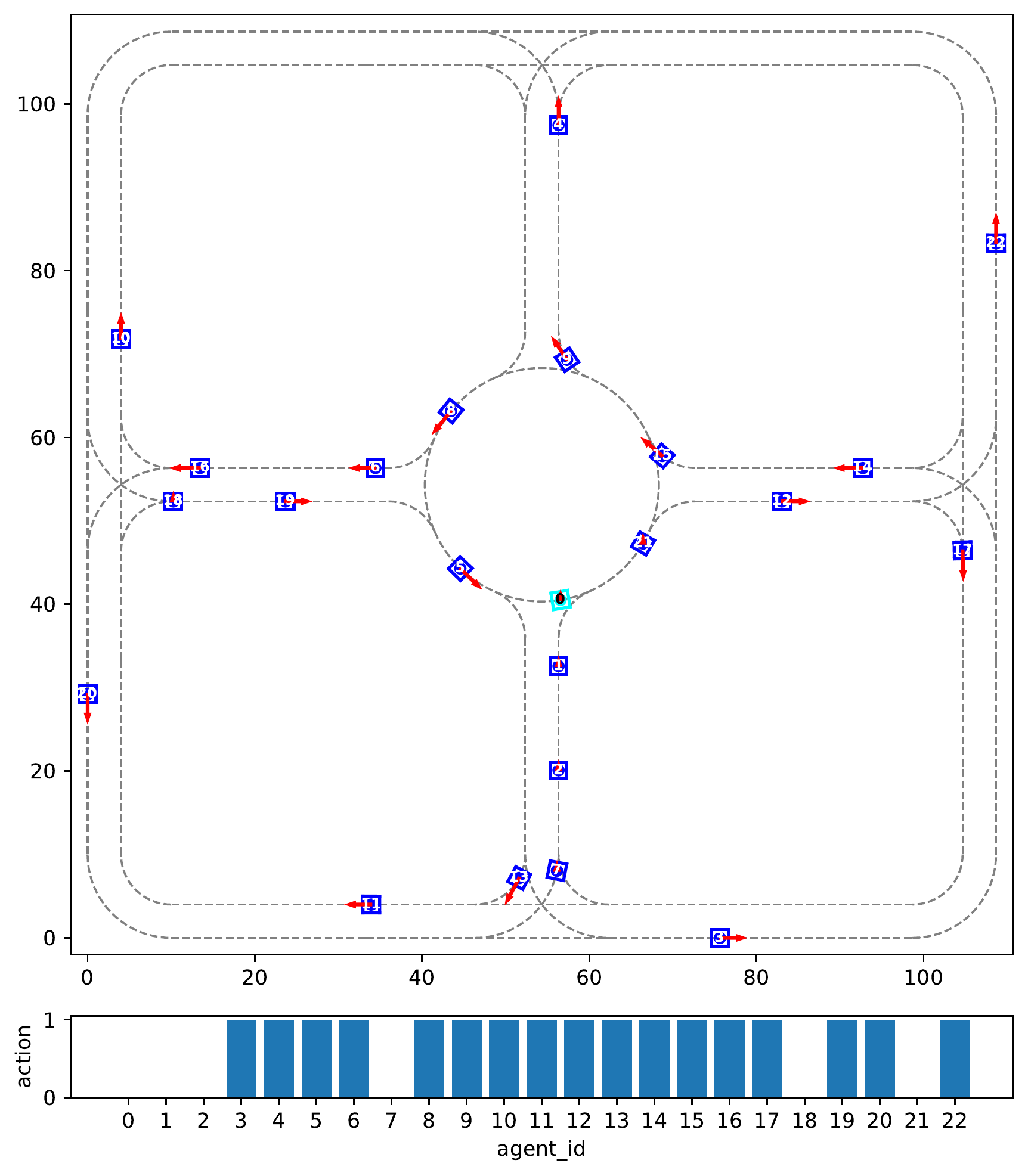}
\caption{t=21}
\label{}
\end{subfigure}
\begin{subfigure}[t]{.3\textwidth}
    \includegraphics[width=\linewidth]{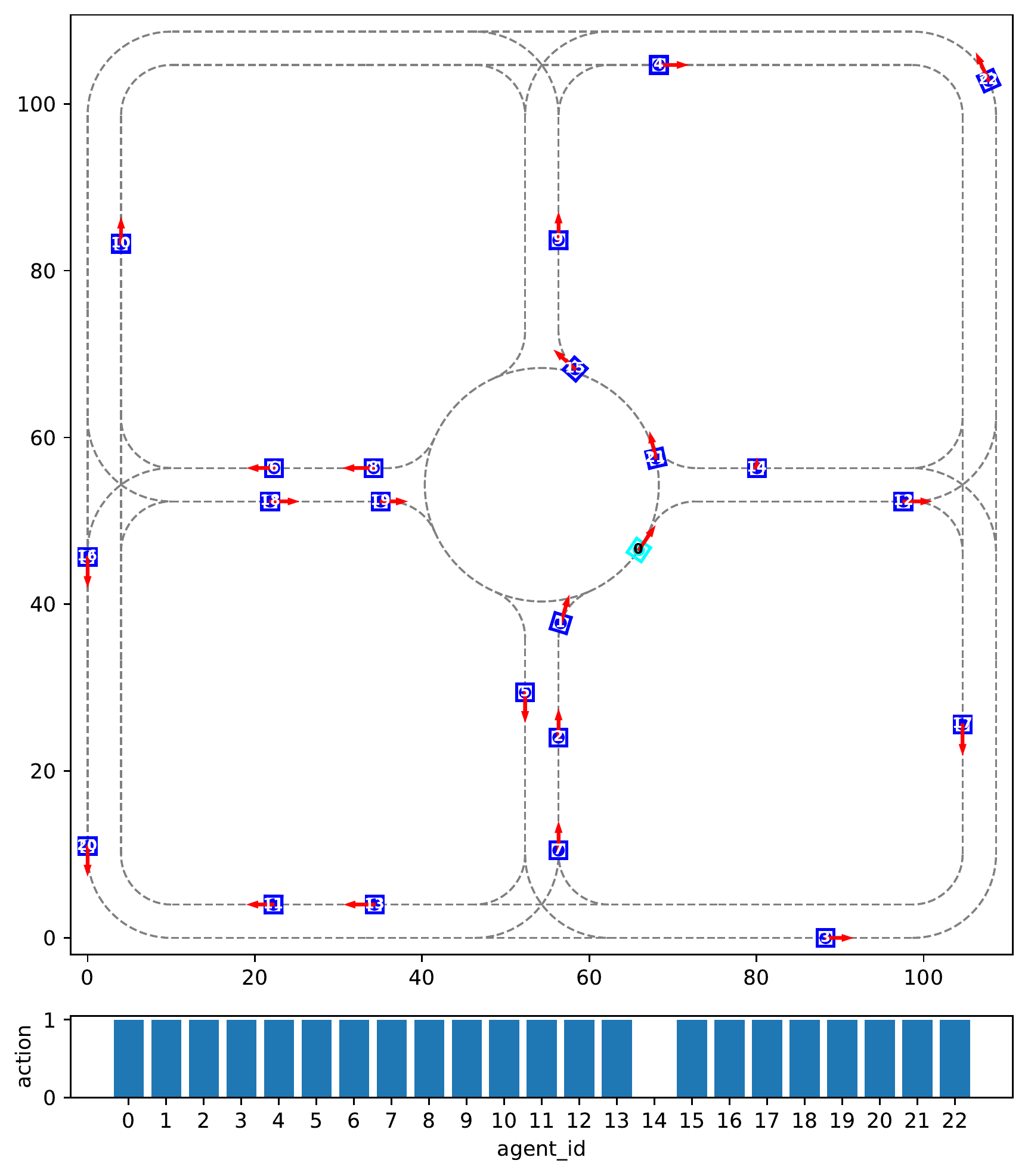}
\caption{t=40}
\label{}
\end{subfigure}
\caption{Illustration of Oracle and MIDAS behavior in car-following episode. Ego is cyan; all other agents are blue. At t=15, ego is approaching a merging point of the roundabout. At t=21, Oracle chooses to stop for front agent at a far distance and blocks the traffic at the intersection. MIDAS chooses to go. At t=40, Oracle chooses to go, since it has information about its own long-term trajectory and knows that ego will turn right. MIDAS doesn't have this information and chooses to stop and keep a distance from the front vehicle.}
\label{fig:car_following_ep}%
\end{figure}

\begin{figure}[h!]%
\centering
\captionsetup[subfigure]{justification=centering}
\begin{subfigure}[t]{.3\textwidth}
    \includegraphics[width=\linewidth]{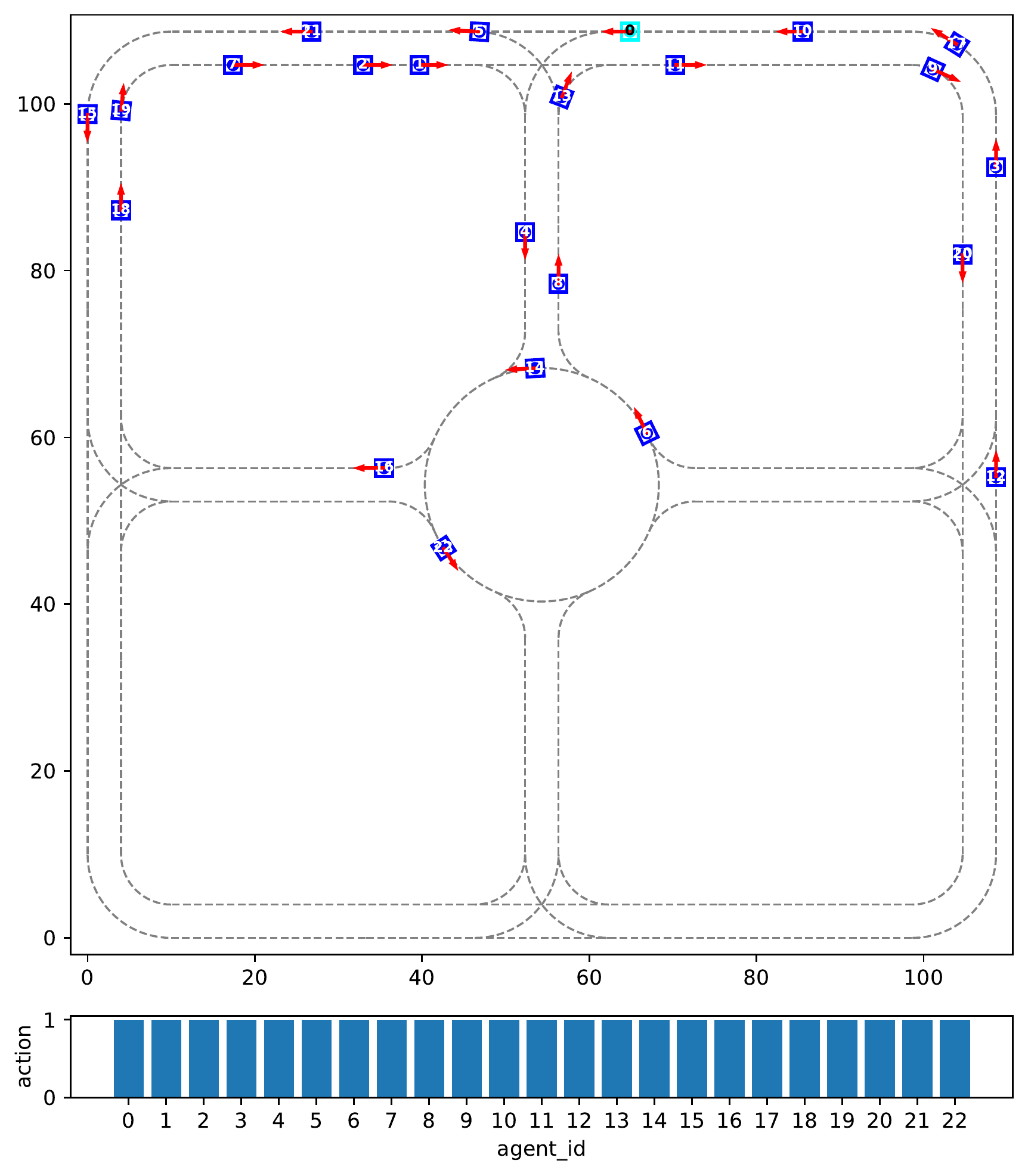}
\caption{t=0}
\label{}
\end{subfigure}
\begin{subfigure}[t]{.3\textwidth}
    \includegraphics[width=\linewidth]{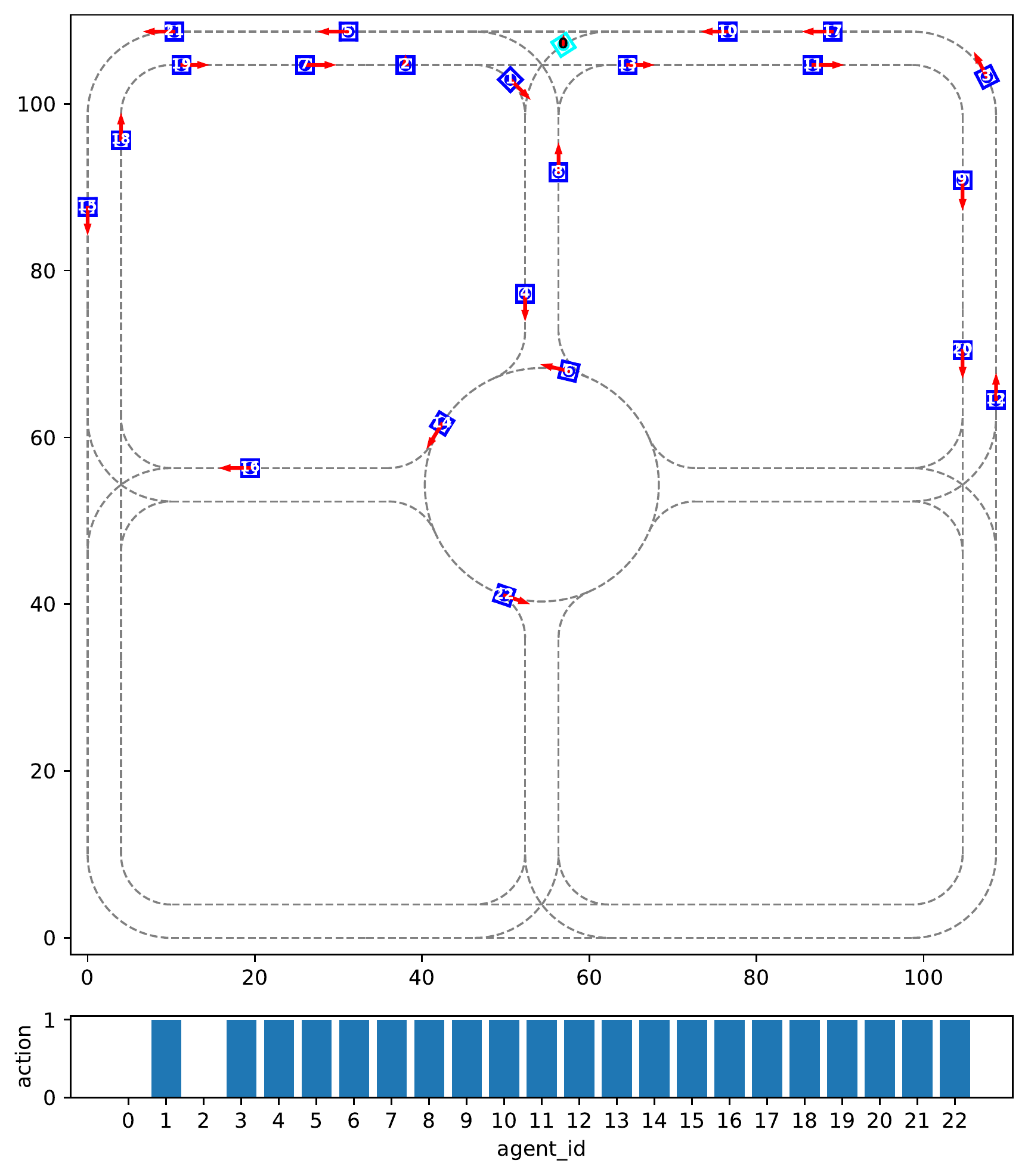}
\caption{t=15}
\label{}
\end{subfigure}
\begin{subfigure}[t]{.3\textwidth}
    \includegraphics[width=\linewidth]{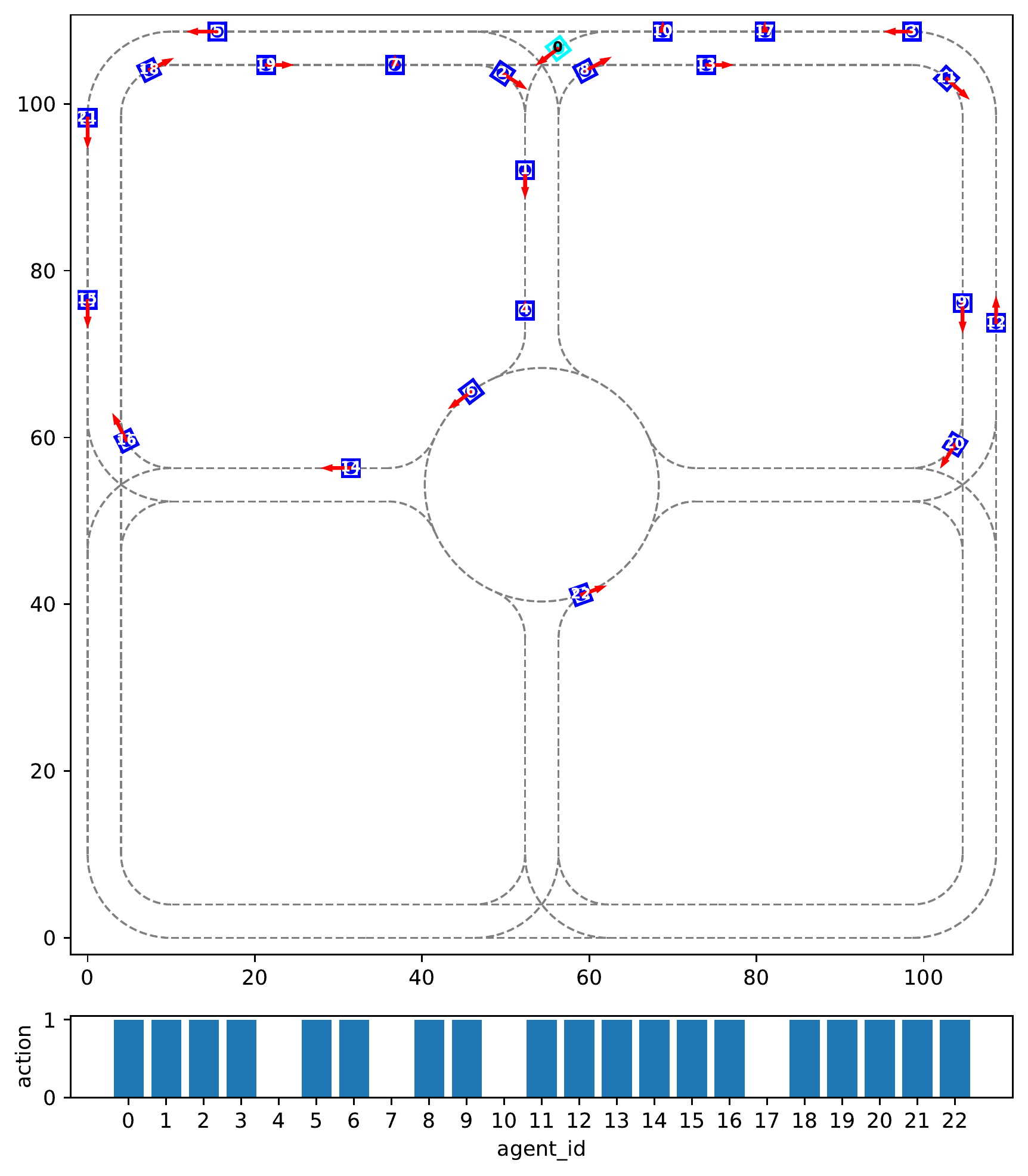}
\caption{t=30}
\label{}
\end{subfigure}

\begin{subfigure}[t]{.3\textwidth}
    \includegraphics[width=\linewidth]{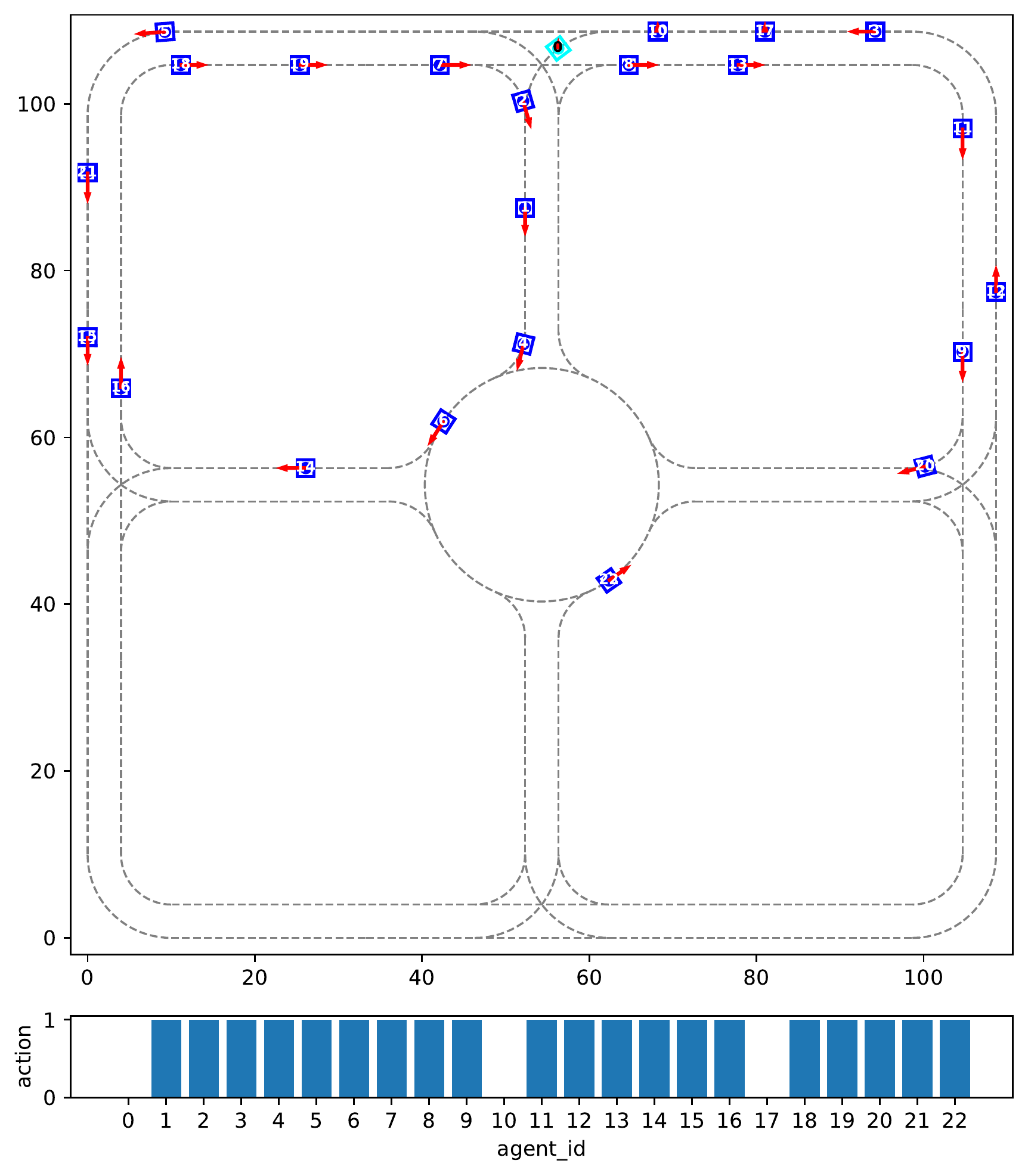}
\caption{t=36}
\label{}
\end{subfigure}
\begin{subfigure}[t]{.3\textwidth}
    \includegraphics[width=\linewidth]{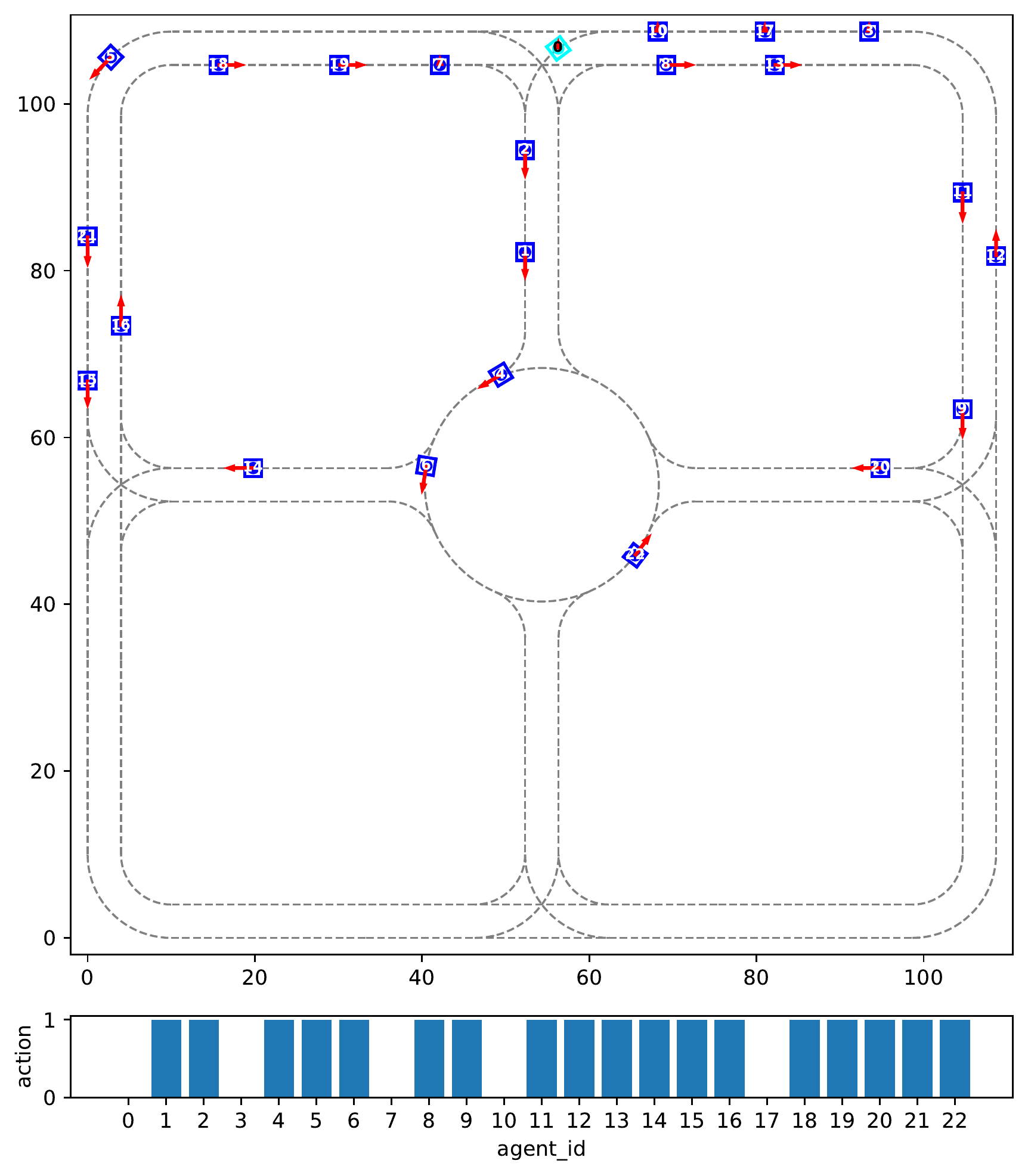}
\caption{t=43}
\label{}
\end{subfigure}
\begin{subfigure}[t]{.3\textwidth}
    \includegraphics[width=\linewidth]{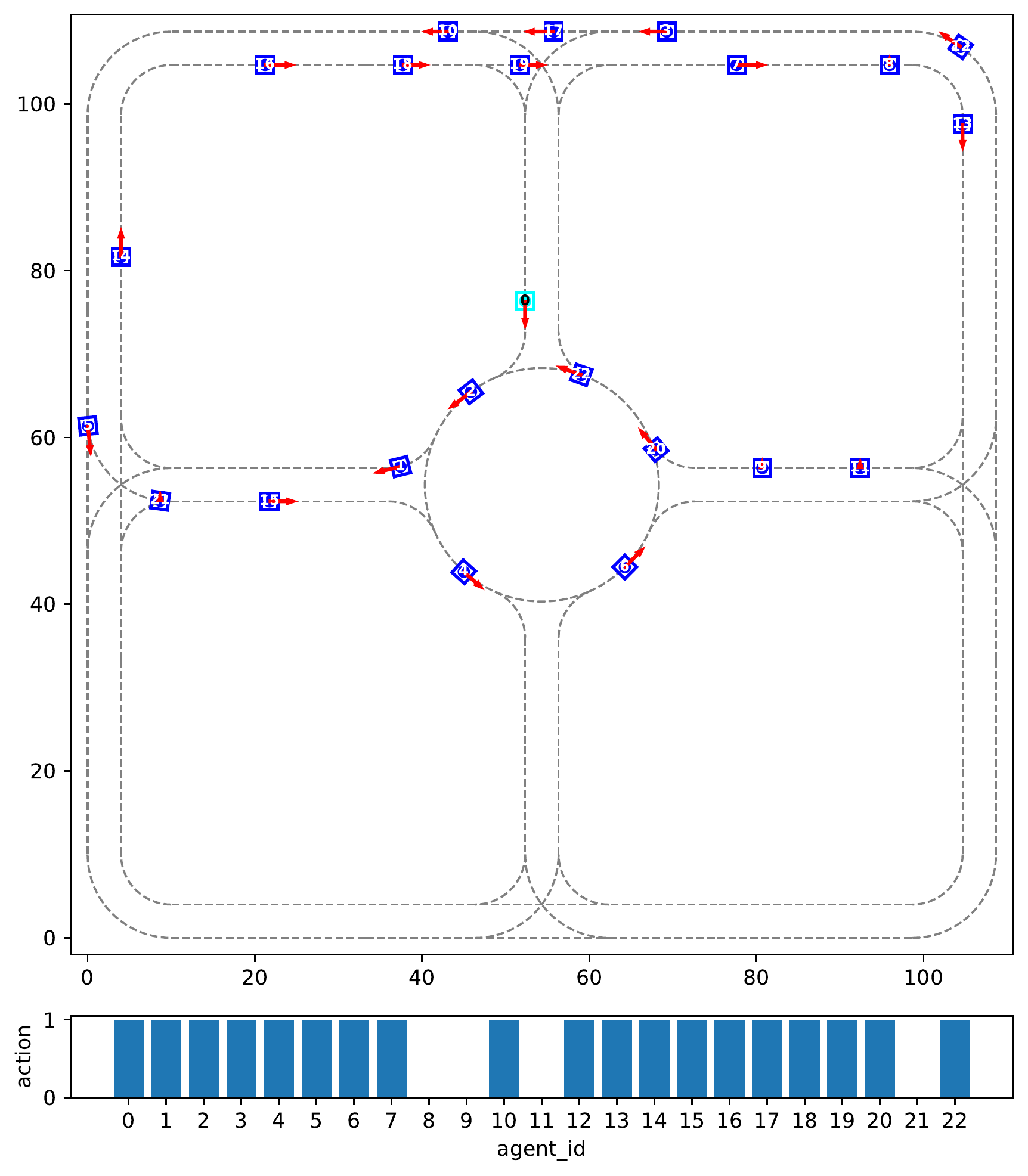}
\caption{t=85}
\label{}
\end{subfigure}
\caption{Illustration of Oracle and MIDAS behavior in left-turn episode. Ego is cyan; all other agents are blue. At t=0, ego is approaching a T-intersection. At t=15, ego stops for a right-turn agent. Oracle stops until t=29, but MIDAS chooses to go shortly after the right turn is finished. At t=30, Oracle wants to go, but was stopped again by a second right-turning agent. Given this state, MIDAS chooses to go at t=36, but Oracle waits until t=43 when there's a big clearance. At t=85, MIDAS stops for an agent in the roundabout, anticipating a right-of-way negotiation, but Oracle keeps going, given global information about tie-braking. This episode shows that MIDAS drives more efficiently but also remains cautious at ambiguous situations, given limited information.}
\label{fig:left_turn_ep}%
\end{figure}

\cref{fig:car_following_ep} and~\cref{fig:left_turn_ep} show the detailed illustration of the car-following and left-turn episodes described in~\cref{ss:results}.
\label{app:typical_episodes}

\clearpage
\subsection{Model Timeout Performance across Ego Driver Types}
\label{app:timeout_rate_across_driver_types}
\begin{figure}[h!]%
    \centering
    \includegraphics[width=0.4\linewidth]{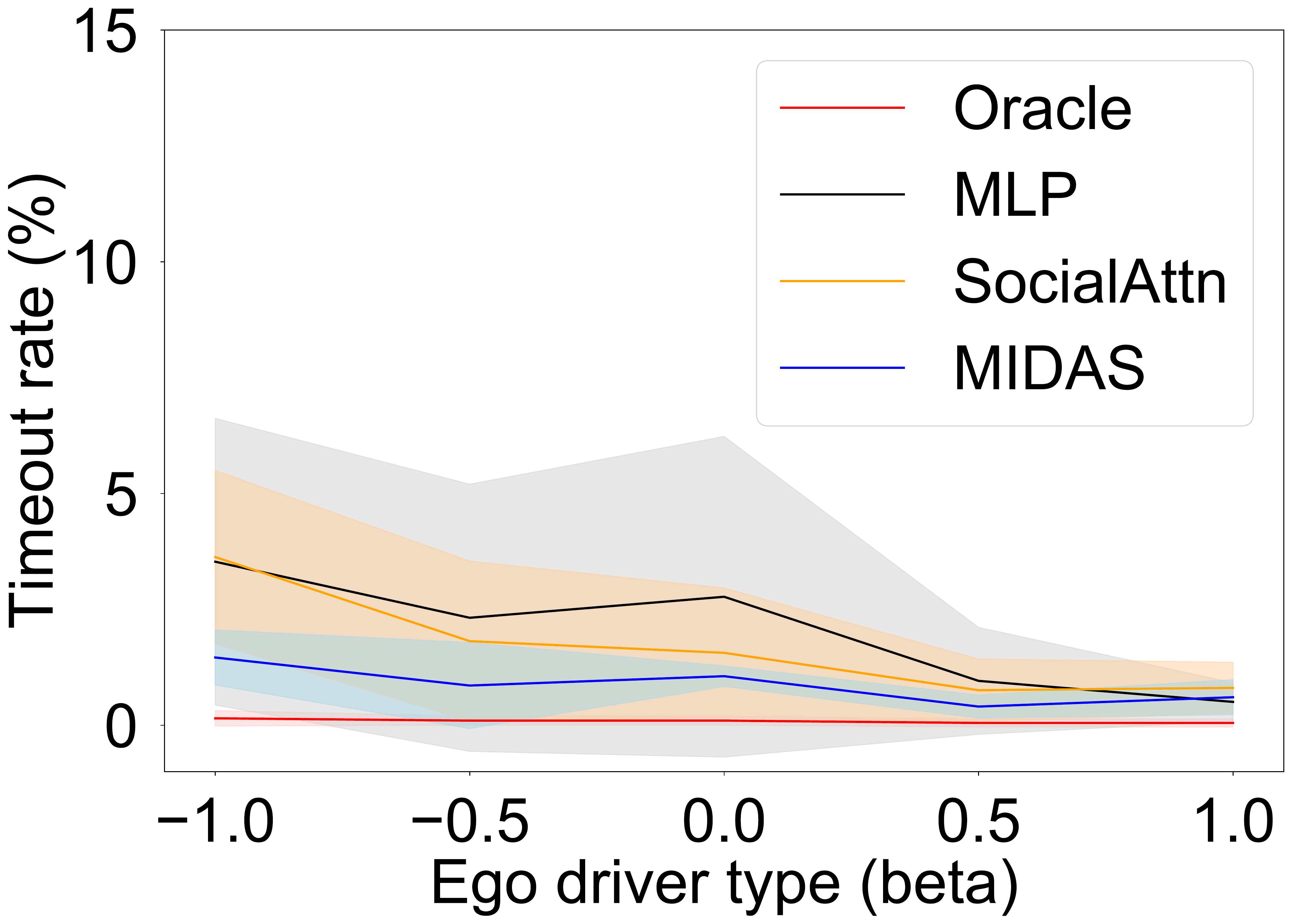}
    \caption{MIDAS has relatively constant timeout rate across different ego driver types, while that of MLP and SocialAttention decreases, accompanied with increasing collision rate, as shown in~\cref{fig:fixed_ego_b1_tr5_eval_set_cls}. SocialAttn refers to SocialAttention.}
    \label{fig:timeout_rate_across_beta}
\end{figure}

\cref{fig:timeout_rate_across_beta} shows the change of timeout rate on the test set across different ego driver types.

\end{document}